\documentclass[10pt,twocolumn,letterpaper]{article}

\usepackage{iccv}
\usepackage{times}
\usepackage{epsfig}
\usepackage{graphicx}
\usepackage{amsmath}
\usepackage{amssymb}
\usepackage{booktabs}
\usepackage{tabularx}
\usepackage{multirow}
\usepackage[normalem]{ulem}
\usepackage{enumitem}
\usepackage{caption}
\usepackage{comment}

\newcommand{\nrsfm}{NRS\textit{f}M\xspace}
\newcommand{\hats}[1]{\skew{4}\hat{#1}}
\newcommand{\ours}{BLiRF\xspace}

\newcommand{\ra}[1]{\renewcommand{\arraystretch}{#1}}

\usepackage[pagebackref=true,breaklinks=true,letterpaper=true,colorlinks,bookmarks=false]{hyperref}

\newcommand{\supprefshort}[1]{Supp.~\ref{#1}}

\iccvfinalcopy 

\def\iccvPaperID{8935} 

\ificcvfinal\fi

\begin{document}

\title{BLiRF: Band Limited Radiance Fields for Dynamic Scene Modeling}

\author{Sameera Ramasinghe\hspace{1em}
Violetta Shevchenko\hspace{1em}
Gil Avraham\hspace{1em}
Anton Van Den Hengel\\
Amazon, Australia\\
}
\twocolumn[{%
\renewcommand\twocolumn[1][]{#1}%
\maketitle
\ificcvfinal\thispagestyle{empty}\fi

\begin{center}
\centering
\includegraphics[width=0.9\textwidth]{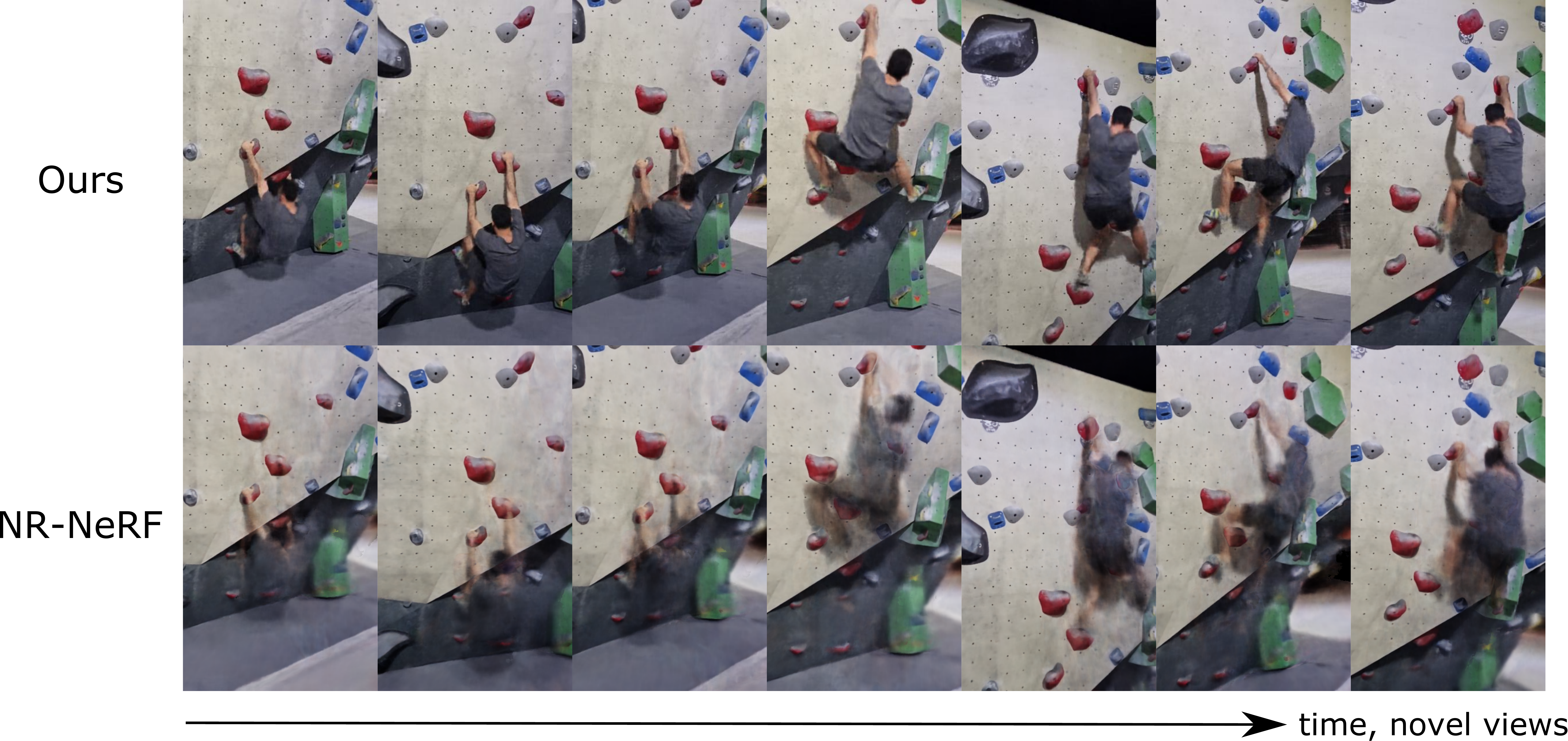}
\captionof{figure}{\small We address the problem of recovering the 3D structure of a dynamic scene given sparse RGB views from a monocular, moving camera. The figure shows a comparison between novel views of a challenging scene with long-range dynamics as synthesized by our proposed model and a competitive baseline. As illustrated, our model is able to capture fine details better and accurately localize the motion compared to  NR-NeRF \cite{tretschk2021non}. We attribute the superior performance of our model to efficient factorization of time and space dynamics that enable incorporating well-defined spatio-temporal  priors, leading to better recovery of complex dynamics.}
\label{fig:teaser}
\end{center}
}]
\begin{abstract}
Inferring the 3D structure of a non-rigid dynamic scene from a single moving camera is an under-constrained problem. 
Inspired by the remarkable progress of  neural radiance fields (NeRFs) in photo-realistic novel view synthesis of static scenes, the approach has naturally been applied to dynamic settings. 
The associated methods heavily rely on implicit neural priors to regularize the problem.  In this work, we take a step back and investigate how current implementations may entail 
deleterious effects including limited expressiveness, entanglement of light and density fields, and sub-optimal motion localization. As a remedy, we propose a bridge between classic non-rigid-structure-from-motion (\nrsfm) and NeRF, enabling the well-studied priors of the former to guide the latter. To this end, we devise a factorisation-based framework that represents the scene as a composition of bandlimited, high-dimensional  signals. We demonstrate  compelling results across complex dynamic scenes that involve changes in lighting, texture and long-range dynamics. Our code and data will be released.



\end{abstract}

 \vspace{-0.5em}
\section{Introduction}
\label{sec:intro}

The problem of scene modeling\cite{kolmogorov2002multi, dyer2001volumetric} is one of the fundamental challenges in  computer vision, and underpins many of the field's most prominent applications including novel view synthesis\cite{avidan1997novel, daribo2010depth}, augmented and virtual reality\cite{azuma1997survey, burdea2003virtual}, and SLAM\cite{grisetti2010tutorial, mur2015orb}. In this vein, Neural Rendering Fields (NeRFs) \cite{mildenhall2021nerf} have recently exhibited remarkable progress in synthesizing photorealistic novel views from sparse 2D images. 


One of the factors underpinning the success of NeRFs is the architectural bias of neural networks. The  (Lipschitz) smoothness of neural networks acts as an implicit \emph{neural prior} for self-regularizing the optimization process, which is otherwise ill posed \cite{zhang2020nerf}. 
Recently, multiple works have extended NeRFs to dynamic settings, 
leveraging the same neural smoothness prior that made NeRFs successful. For modeling the evolution of scene  geometry over time, these works have primarily resorted to using ray deformation paradigms, which parameterize rays cast from the camera as functions of time \cite{pumarola2021d, tretschk2021non, park2021nerfies, li2021neural, gao2021dynamic, xian2021space, park2021hypernerf}. Although these approaches have yielded impressive results, we show that their over-reliance on implicit neural priors gives rise to fundamental problems; \textit{a}) Dependency on a canonical frame which harms modeling long range motion, \textit{b}) Entanglement of the light and density fields \textit{c}) Limited expressiveness due to network bottlenecks, and \textit{d}) substandard localization of motion due to the difference in the spectral properties of space and time, \textit{i.e.}, space typically consists of sharp/high-frequency details, whereas temporal dynamics are generally smooth and continuous.

To overcome the above drawbacks, we propose a theoretical framework that enables efficient integration of implicit neural priors and well-defined explicit priors.  On this basis we also propose a set of explicit priors inspired by non-rigid-structure-from-motion (\nrsfm). 
Over many years the \nrsfm literature developed a series of explicit priors applicable to the otherwise ill-posed problem of recovering the 3D structure of deformable objects and scenes from 2D point correspondences. 
The performance of \nrsfm methods depends critically on the alignment of these priors with the deformation in question. Thus, our work can be considered as a bridge between NeRF and \nrsfm.

In particular, we model the light and density fields of a 3D scene as bandlimited, high-dimensional signals. This  standpoint enables complete factorization of spatio-temporal dynamics, allowing us to inject explicit priors on the time and space dynamics independently. To demonstrate the practical utility of our framework, we offer an example implementation that enforces 1) a low-rank constraint on the shape space, along with 
2) a neural prior over the frequency domain
and 3) a union-of-subspaces prior 
on the deformation of a shape over time.
We show that the strong regularization effects of these priors enable our model to reconstruct long-range dynamics and localize motion accurately. Further, our model does not rely on complex optimization procedures \cite{pumarola2021d, li2021neural, park2021nerfies, yoon2020novel, park2021hypernerf} or multiple explicit loss regularizations \cite{tretschk2021non, gao2021dynamic, park2021nerfies, li2021neural, wang2021neural} that are common in existing dynamic NeRF works, indicating the stability of our formulation. Finally, our implementation efficiently disentangles light and density fields, allowing the model to capture challenging scenes with dynamic lighting and textures. Our contributions are summarized as follows.

\begin{itemize}[topsep=-2pt,itemsep=1pt]
    \item We show that existing extensions of NeRF to dynamic scenes suffer from critical drawbacks, primarily due to their over-reliance on implicit neural priors.
    
    \item We propose a novel framework for modeling dynamic 3D scenes that overcomes the above drawbacks by formulating radiance fields as bandlimited signals. 
    
    \item We empirically validate the efficacy of our framework by demonstrating better modeling of long-range dynamics, motion localization, and light/texture changes. Our model only takes around $3$ hours per scene to train (more than $10$ times faster than NR-NeRF, D-NeRF, and HyperNeRF), and does not require complex loss regularizers or optimization procedures. 
\end{itemize}

 \vspace{-0.5em}
\section{Related Work}
\label{sec:related}
 \vspace{-0.5em}

\noindent \textbf{\nrsfm.} The problem of \nrsfm focuses on modeling the 3D structure of a set of sparse points on the basis of their 2D projections into a set of images.  The non-rigid aspect of the problem refers to fact that the points might move relative to each other over time. The \nrsfm problem is inherently ill-defined, and various additional priors have been explored. 


Breger \etal~\cite{bregler2000recovering}, in their seminal work, argued that \nrsfm could be solved using a finite number of low-rank shape-basis functions\cite{garg2013dense}. Later, Torrensani \etal \cite{torresani2001tracking} modeled the coefficients of the shape-basis as a linear dynamical system. In contrast, Rabaud \etal \cite{rabaud2008re} proposed to learn a smooth manifold of shape configurations, and Gotardo \etal \cite{gotardo2011kernel} explored non-linear shape models using kernels. More recently, Agudo \etal \cite{agudo2017dust} imposed a union-of-subspace prior to constrain the shape deformations. Considering trajectory-based priors, Akhter \etal~\cite{akhter2008nonrigid} demonstrated that instead of decomposing the shape deformation over time with basis functions, the trajectory of measurements could be formulated as DCT basis functions. In the same spirit, \cite{zhu2013convolutional} exploited the convolutional structure of the trajectories. Multiple works ~\cite{kumar2017spatio, agudo2017dust, zhu2014complex, zappella2013joint} showed that frames could be clustered to restrict trajectories within low-dimensional subspaces. This closely aligns with the manifold prior we propose in Sec.~\ref{sec:manifold}. Multiple works have also sought to explicitly regularize trajectories by minimizing their response to high-pass filters \cite{valmadre2012general}, injecting rigid key-frames~\cite{zhu20113d}, enforcing sparsity priors \cite{salzmann2011physically}, and considering articulated motion~\cite{park20113d}. 


\noindent \textbf{Dynamic NeRF.} Inspired by the success of NeRF,  many  dynamic neural radiance field models have been developed, primarily using the concept of ray deformation  \cite{pumarola2021d, tretschk2021non, park2021nerfies, li2021neural, gao2021dynamic, park2021hypernerf}.  D-NeRF\cite{pumarola2021d} was first to propose a general framework which learns a per-point displacement against a canonical location to model ray deformation. Both\cite{tretschk2021non, gao2021dynamic} further introduced a constraint to model the foreground and background separately, thus allowing  quicker convergence and a better-constrained search space. \cite{gao2021dynamic} introduced a method to disambiguate self-occlusions that hinders the performance of these approaches.  Nerfies\cite{park2021nerfies} achieves remarkable results on novel views synthesis of dynamic scenes by incorporating elastic regularization, but specifically target self-portraits. Finally, several other NeRF extensions have also been proposed that require depth estimates \cite{xian2021space}, optical flow \cite{wang2021neural},  foreground masks~\cite{johnson2022unbiased, gao2021dynamic}, meshes~\cite{xu2022deforming}, or assume that dynamic objects are distractors to be removed \cite{martin2021nerf}.

\section{Revisiting ray deformation networks}
\label{sec:ray_bending}




Extending NeRF to dynamic scenes requires representing the scene as a continuous function with 6D inputs $(x,y,z,\theta, \phi, t)$, where $t$ is the time and $(\theta, \phi)$ is the viewing direction. However, it has been shown empirically~\cite{pumarola2021d} that employing a single MLP to learn a mapping from 6D inputs to density and color fields yields sub-optimal results. Hence, existing works decompose the aforementioned task into two modules \cite{pumarola2021d, tretschk2021non, park2021nerfies, li2021neural, gao2021dynamic, park2021hypernerf, TiNeuVox}: 1) the first MLP learns a warping field of 3D points $(\Delta x, \Delta y, \Delta z)$ sampled along the rays with respect to a canonical setting;  2) the second module then acts similarly to the original NeRF formulation, regressing the density and light fields given the warped  samples along the rays $(x + \Delta x, y + \Delta y, z + \Delta z)$. Since the warping is applied to points sampled along the ray, this formulation is interpreted as deforming the rays as a function of time. Further, note that this assumes that objects do not enter or leave the scene, and that lighting/texture is consistent.
However, we notice that existing implementations of this framework do not adhere to these constraints (see \supprefshort{suppsec:ray_deformation}). Specifically, we show that such networks can indeed model light and density changes separately (to an extent), which is infeasible with a model that only learns ray deformations (see Fig.~\ref{fig:real} and Fig.~\ref{fig:synthetic}). However, to avoid confusion, we will keep referring to this class of models as ray deformation networks. Next, we discuss several critical limitations thereof.

\subsection{Limitations of ray deformation networks}
\vspace{-5pt}
\label{subsec:limitations}

In this section, we present a brief exposition of the limitations entailed in the ray deformation approach. For an extended analysis, refer to \supprefshort{suppsec:ray_deformation}.

\noindent \textbf{Dependency on a canonical frame: }Ray deformation networks require choosing a canonical frame, and most models choose the frame at $t=0$ to this end. The particular choice of frame can significantly harm model performance in cases where 1) objects or the camera exhibit long-range translations, and 2) new objects appear in subsequent frames of the video. The canonical frame thus needs to provide a form of average scene representation where all future information is present. This becomes increasingly infeasible as the scene becomes more complex. On the other hand, the ray deformation model also needs to preserve continuity; the model output at $(t=\delta t)$ needs to embody a smooth transition of the canonical scene at $t=0$, which can be impractical if the scene comprises abundant future information. In contrast, our framework does not use ray deformation and thus does not depend on a canonical scene.

\noindent \textbf{Entanglement of light and density fields: } Although ray deformation networks are able to deform the light and density fields, they are still highly entangled. More precisely, it can be shown that in order to achieve complete disentanglement of the light and density fields, the network needs to preserve a specific block-diagonal Jacobian structure in one of the hidden layers, which is an extremely restrictive requirement (\supprefshort{suppsec:ray_deformation}). In comparison, our framework achieves complete disentanglement by design,  modeling the light and density fields independently.

\noindent \textbf{Limited expressiveness: } Ray deformation networks comprise a bottleneck of dimension three. Therefore, each of the density and light fields modeled by this network becomes a three dimensional manifold. They cannot thus encode complex dynamics that need to be parameterized by four variables $(x,y,z,t)$ (\supprefshort{suppsec:ray_deformation}).

\noindent \textbf{Substandard separation of background and motion: } Ray deformation networks model the warp field using a single MLP. However, this is a substandard design choice since the space and time variations have different spectral properties. Natural scenes often exhibit high-frequency spatial characteristics such as fine-grained surface details for example, but temporal changes are generally smoother. Therefore, using an MLP with a particular bandwidth for learning both spatial and time variations together leads to sub-optimal reconstructions. One way to overcome this problem is to use separate MLPs for time and space dynamics, and control their bandwidth via positional encodings.  This strategy requires involved optimization procedures that demand careful coarse-to-fine hyperparameter annealing that depends on the dataset, and incur longer training times ($64$ hours on $4$ TPU v4s vs ours ($3$ hours on a single V100)) \cite{park2021hypernerf}. In contrast, our framework enables much more elegant factorization of space and time dynamics by modeling the scene as bandlimited signals, allowing better separation of static and dynamic regions.

\vspace{-1.em}
\section{Our framework}
\label{sec:method}
\vspace{-0.5em}
\begin{figure*}[t]
\centering
\includegraphics[width=0.7\textwidth]{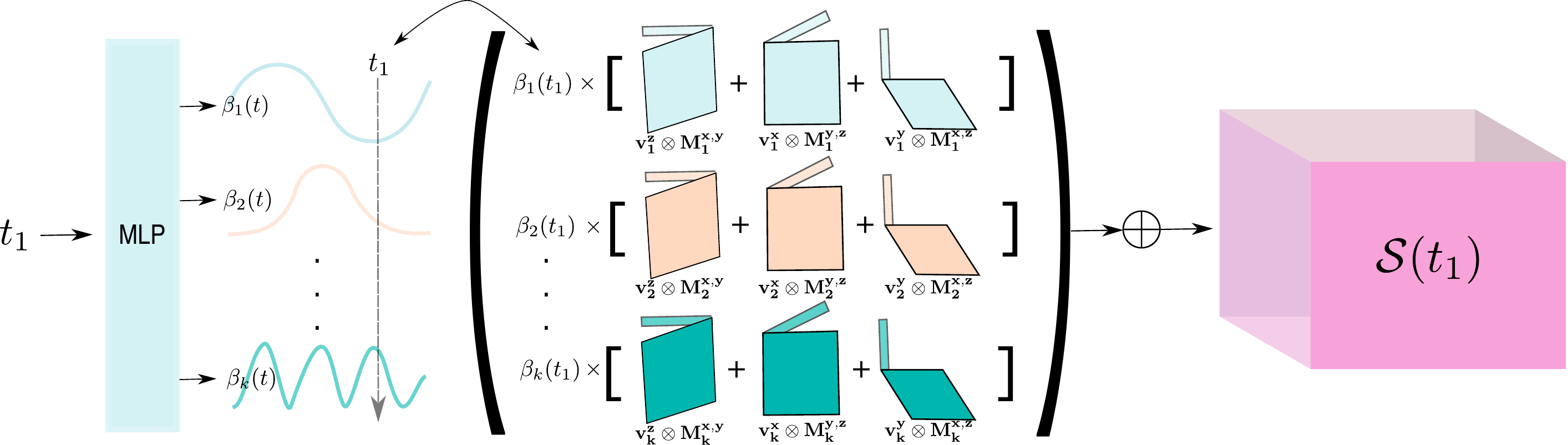}
\caption{\small \textbf{The proposed implementation of our framework.} We treat the light and density fields as bandlimited, high-dimensional signals (only a single field is shown in the figure). The time evolution of each 3D point $(x,y,z)$ of the field is modeled as a finite linear combination of time-basis functions $\{\beta_j(t)\}$. The coefficients of the $\{\beta_j(t)\}$  are decomposed into outer products between learnable matrices ($\mathbf{M}$) and vectors ($\mathbf{v}$). This decomposition is inspired  \cite{chen2022tensorf}. Our formulation allows efficient factorization of time and space dynamics, leading to high-quality reconstructions of complex dynamics, along with faster convergence. } 
\label{fig:arch}
\end{figure*}

Consider a set of 2D projections $\{I(t_n)\}_{n=1}^N$ of a 3D scene captured from a moving camera. For brevity, we drop the dependency on the camera poses from the notation. Without the loss of generality, we assume that the scene is bounded within a cube with side length $D$. We begin by observing that  there exists a latent density and color field corresponding to each $I(t_n)$ that can be discretized into a cubic grid of $D^3$ nodes. Then, rewriting the latent states of either field in the matrix form, gives





\begin{equation}
    \resizebox{\columnwidth}{!}{%
        $\mathbf{S} =  \begin{bmatrix}
        s(t_1, \mathbf{x}_1) & s(t_1, \mathbf{x}_2)  & \dots &  s(t_1, \mathbf{x}_{D^3})  \\
        \vdots & \vdots & \ddots & \vdots\\
        s(t_N, \mathbf{x}_1) & s(t_N, \mathbf{x}_2)  & \dots &  s(t_N, \mathbf{x}_{D^3})  \\
        \end{bmatrix}_{N \times D^3}$%
    }
\end{equation}

%
%
%
where $\mathbf{x}_i \in \mathbb{R}^3$ and $s(t_n, \mathbf{x}_i)$ can be either the density or the color values emitted from $\mathbf{x}_i$ at time $t_n$. Note that $s(\cdot, \cdot)$ can be either a scalar valued function for density and a vector valued function for color. To avoid cluttered notation, we consider the scalar valued case for the following derivations. However, our analysis holds true for the vector valued case as well;
Let $\mathrm{rank}(\mathbf{S}) = K$. Then, there exist $K$ basis vectors, each with dimension $D^3$, that can perfectly reconstruct (memorize) $\mathbf{S}$. More precisely, in this case, each row of $\mathbf{S}$ can be reconstructed as
\begin{equation}
      \mathbf{S} (t_n, \cdot ) = \sum_{j=1}^K a_{j,n} \; \hat{\boldsymbol{\alpha}}_j,
\end{equation}
where $\mathbf{S} (t_n, \cdot )$ is the $n^{th}$ row of $\mathbf{S}$, $\{ \hat{\boldsymbol{\alpha}}_j \}_{j=1}^K$ are basis vectors of dimension $D^3$, and $\{ a_{j,n} \}$ are scalar coefficients. Intuitively, each row of $\mathbf{S}$ corresponds to a snapshot of the field (in space) at a particular time instant.  On the contrary, each column of $\mathbf{S}$ is a representation evolution of a particular point  $\mathbf{x}$ over time. We note an interesting duality here; since the dimension of the row space and the column space of $\mathbf{S}$ are equal, it should be possible to reconstruct the evolution of the density/color value of each  position $\mathbf{x}$ over time using $K$  basis vectors. Thus, we model the time evolution of each point as
 \vspace{-0.5em}
\begin{equation}
      \mathbf{S}(\mathbf{x}_i, \cdot) = \sum_{j=1}^K b_{j,i} \; \hats{\boldsymbol{\beta}}_j,
      \label{eq:time-memorization}
\end{equation}
where $\mathbf{S}(\mathbf{x}_i, \cdot)$ is the $i^{th}$ column of $\mathbf{S}$,  $\{\hats{\boldsymbol{\beta}}_j\}_{j=1}^{K} \in \mathbb{R}^N$ are basis vectors, and $\{ b_{j,i} \}$ are scalars.
This change of perception is crucial for generalizing to unseen time instances  and obtaining a space-time factorization, as we  shall discuss next.
Using Eq.~\ref{eq:time-memorization}, the value of color/density of a point $\mathbf{x}$ at a particular continuous time instance $t$ can be obtained as




\begin{equation}
    \mathbf{S}(\mathbf{x}, t) = \sum_{j=1}^K \Tilde{b}_{j}(\mathbf{x}) \; \psi_j(t),
    \label{eq:delta}
\end{equation}

where $\psi_j(t) = \sum_{n=1}^N \hats{\boldsymbol{\beta}}_{j,n}\mathbf{\delta}(t-t_n)$,  $\mathbf{\delta}(\cdot)$ is the Dirac delta function, and $\hats{\boldsymbol{\beta}}_{j,n}$ is the $n^{th}$ element of $\hats{\boldsymbol{\beta}}_j$. $\Tilde{b}_{j}(\mathbf{x})$ is a scalar valued function. A key problem associated with Eq.~\ref{eq:delta} is that $\psi(t)$ is an infinite bandwidth function because $\delta$ is of infinite bandwidth. As a consequence, $\mathbf{S}(\mathbf{x}, t)$ also becomes an infinite bandwidth function. Equivalently, an infinite number of time-sample points are required to reconstruct the continuous signal $\mathbf{S}(\mathbf{x}, t) $\footnote{Recall that in order to reconstruct a continuous signal as a linear combination of shifted Dirac delta functions, an infinite number of sampling points are needed.}.  Recall, however, that, in practice, only a sparse, finite set of 2D observations $\{I(t_n)\}_{n=1}^N$ are at our disposal. Therefore, the infinite bandwidth representation of Eq~\ref{eq:delta} is not ideal for obtaining a  $\mathbf{S}(\mathbf{x}, t)$ that can be queried at arbitrary continuous time instances. Therefore, we replace $\{\psi_j\}$ with a set of bandlimited scalar-valued functions $\{\beta_j(\cdot)\}$. 

\begin{equation}
    \mathbf{S}(\mathbf{x},t) = \sum_{j = 1}^K  \Tilde{b}_{j}(\mathbf{x})\; \beta_j(t).
    \label{eq:bandlimited}
\end{equation}


From a signal processing perspective, this can also be considered as reconstructing a signal from discrete samples using a linear combination of bandlimited basis functions. Observe that from this perspective, the $\psi_j(t)$'s can be considered  discrete samples of the continuous functions $\beta_j(t)$. Consequently, $\mathbf{S}(\mathbf{x},t)$ also becomes a bandlimited function (a linear combination of bandlimited functions is bandlimited).  Note that now we have also obtained a  factorization of time and spatial dynamics that will allow us to impose priors on time and space independently. In Sec.~\ref{sec:implementation}, we present an implementation of the proposed framework. In this implementation, we inject a low-rank prior on space, along with smoothness and compact manifold priors on time. It is worth to note that our framework is generic enough to support alternative implementations and more complex priors, which we leave to future explorations.


\section{Implementation}
\vspace{-0.5em}
\label{sec:implementation}
Leveraging the factorization we achieved in Eq.~\ref{eq:bandlimited}, we can formulate the entire 3D field volume as a time-dependent higher dimensional signal, that can be decomposed into a linear combination of 3D tensors $\mathbf{\mathcal{A}}^{xyz}_j \in \mathbb{R}^{D \times D \times D}$:
 \vspace{-0.5em}
\begin{equation}
    \mathcal{S}(t) = \sum_{j=1}^K\beta_j(t) \mathbf{\mathcal{A}}^{xyz}_j,
\end{equation}
where $\mathcal{S}(t) \in \mathbb{R}^{D \times D \times D}$ is the state of the field at time $t$. Note that we adopt the tensor notation here where the superscripts denote the dimensions, \textit{i.e.}, $x = 1, \dots D, y = 1, \dots, D$, and $z = 1, \dots, D$.
To regularize the spatial variations, we employ a low-rank constraint on $\mathbf{\mathcal{A}}_j$ as,
 \vspace{-0.5em}
\begin{equation}
  \mathcal{S}(t) = \sum_{j=1}^K\beta_j(t) ( \mathbf{v}^z_j \otimes \mathbf{M}_j^{xy} + \mathbf{v}^x_j \otimes \mathbf{M}_j^{yz} + \mathbf{v}^y_j \otimes \mathbf{M}_j^{xz}),
\label{eq:decompose}
\end{equation}
where $\mathbf{v}_j \in \mathbb{R}^D$ and $\mathbf{M}_j \in \mathbb{R}^{D\times D}$ are one- and two-dimensional tensors, respectively, and $\otimes$ is the outer product. The above choice of factorization is inspired by the \emph{VM-decomposition} proposed in \cite{chen2022tensorf}. This factorization accomplishes two goals: 1) enforcing a low rank constraint on the spatial variations of the field, and 2) significantly reducing the size of the model and the number of trainable parameters. We note that such low-rank priors have been widely employed in the \nrsfm literature for the same purpose \cite{torresani2001tracking, torresani2003learning, rabaud2008re}.

\subsection{Neural trajectory basis}
\vspace{-0.5em}
In theory, it is possible to use any class of bandlimited functions that form a complete basis in $L^2(\mathbb{R}, dt)$ as $\{\beta_j(t)\}$. Popular choices include the DCT, Fourier, and Bernstein basis functions, among many others. 
Nonetheless, we use neural networks to parameterize our basis functions, leveraging the implicit architectural smoothness constraint built into them. We label these basis functions as the \emph{neural trajectory basis}. The neural trajectory basis presents an important implicit prior to our model, that the field values should evolve smoothly. We also empirically note that neural basis functions are naturally more expressive and adaptive as they are learned end-to-end, as opposed to other choices (see Table~\ref{tab:abl-basis}). Expressiveness is crucial, as it is desirable to model the dynamics of each point with a minimal number of basis functions. Thus, we compute  $\{\beta_j(t)\}$ via an MLP $\mathcal{F}(t): \mathbb{R} \to \mathbb{R}^K$ as,
 \vspace{-0.5em}
 \begin{equation}
     \mathcal{F}(t) = [\beta_1(t), \beta_2(t), \dots, \beta_K(t)].
 \end{equation}
 
 We also show that the smoothness prior embedded into the neural trajectory basis closely aligns with Valmadre \etal  \cite{valmadre2012general}, where they showed that, in \nrsfm, a trajectory's response to high-pass filters should be minimal. We validate that neural trajectory basis exhibits this property in \supprefshort{suppsec:neural_basis}. Overall architecture is depicted in Fig.~\ref{fig:arch}.

 \subsection{Manifold Regularization}
 \vspace{-0.5em}
 \label{sec:manifold}
 Multiple works in \nrsfm have explored restricting the subspace of dynamics in order to obtain better reconstructions. The high-level objective is to temporally cluster the motion in order to restrict similar dynamics to a low-dimensional subspace \cite{kumar2017spatio, agudo2017dust, zhu2014complex, zappella2013joint}. We observed that such a constraint can improve our reconstructions also.
More formally, we empirically asserted that better results are obtained by locally restricting the dimension of the submanifold that $\mathcal{S}(t)$ is immersed in.
Instead of clustering the motion across the entire sequence, we assume that dynamics are locally compact: movements that occur over a small time period can be described using a smaller subspace. To enforce this constraint, we adopt the following procedure.
 
Observe that  $\mathcal{S}(t)$ is a $1$-dimensional manifold embedded in a $D^3$-dimensional space (its local coordinate chart is a compact subspace in $\mathbb{R}$).  Further, at any given time $t$, $\mathcal{S}(t)$ is a linear combination of $K$ points $\{ \mathbf{v}^z_j \otimes \mathbf{M}_j^{xy} + \mathbf{v}^x_j \otimes \mathbf{M}_j^{yz} + \mathbf{v}^y_j \otimes \mathbf{M}_j^{xz} \}_{j=1}^K \in \mathbb{R}^{D \times D \times D}$. Therefore, 
$\mathcal{S}(t)$ is a submanifold of $\mathbb{R}^K$. 

Now, let  
$\mathbf{P}_j^{xyz} = (\mathbf{v}_j^{z} \otimes \mathbf{M}_j^{xy} + \mathbf{v}^x_j \otimes \mathbf{M}_j^{yz} + \mathbf{v}^y_j \otimes \mathbf{M}_j^{xz})
$. Suppose the dimension of the local submanifold we need is $W$, such that $K = dW$ for some integer $d$. Then, we define the 4D tensor $\mathbf{Q}_{j:j+W}^{xyzu} \in \mathbb{R}^{D \times D \times D \times W}$ such that 
$
\mathbf{Q}_{j:j+W}^{xyzu} = \{\mathbf{P}_u^{xyz}\}_{u=j}^{j+W}.
$
 Next, we obtain
 \vspace{-0.5em}
 \begin{equation}
    \resizebox{\columnwidth}{!}{%
        $\mathbf{\Tilde{Q}}^{xyzu} (t) =  \sum\limits_{n=0}^{d-1} \mathbf{Q}_{(nW + 1):W(n+1)}^{xyzu} \odot \mathrm{sinc}  \big( (d-1)(t - \frac{n}{(d-1)})  \big),$%
    }
 \end{equation}
where $\odot$ represents element-wise multiplication, and 
\[\mathrm{sinc}(r) =  
\begin{cases}
   1, & \text{if } r = 0\\
    \frac{\mathrm{sin}(r)}{r},              & \text{otherwise}
\end{cases}.
\]
The choice of the sinc$(\cdot)$ function here is not arbitrary, and is crucial for the smooth transition between submanifolds as the time progresses. More precisely, the sinc interpolation ensures that no  frequencies higher than $(d-1)/2$ can be presented in $\mathbf{\Tilde{Q}}^{xyzu} (t)$ along the temporal dimension. Finally, we can obtain the regularized field as
 \vspace{-0.5em}
\begin{equation}
    \mathcal{\Tilde{S}}(t) = \sum_{u=1}^W\beta_u(t)\mathbf{\Tilde{Q}}^{xyzu} (t).
    \label{eq:manifold}
\end{equation}

From a strict theoretical perspective, one can argue that Eq.~\ref{eq:manifold} violates the time and space factorization we obtained in Eq.~\ref{eq:decompose}. However, in practice, the sinc interpolation ensures that $\mathbf{\Tilde{Q}}^{xyzu} (t)$ is locally almost constant as long as we choose $d$ to be suitably small, as $\mathbf{\Tilde{Q}}^{xyzu} (t)$ cannot then have higher frequencies than $(d-1)/2$. Further, Eq.~\ref{eq:manifold} ensures that $\mathcal{\Tilde{S}}(t)$ can only locally traverse within an $\mathbb{R}^W$ subspace where $W < K$, which is a more regularized setting than Eq.~\ref{eq:decompose}, where $\mathcal{S}(t)$ is allowed to traverse within an $\mathbb{R}^K$ subspace.

 \section{Training}
 \vspace{-0.5em}
 
 Let $\sigma(\mathbf{x},t), c(\mathbf{x},t)$ be density and light values, queried at position $\mathbf{x}$ at time $t$ (obtained via Eq.\ref{eq:manifold}). To compute the above values at an arbitrary continuous position $\mathbf{x}$, we tri-linearly interpolate the grids. We perform volumetric rendering as in \cite{mildenhall2021nerf} to predict pixel colors $\Tilde{p}$ for each training image (see \supprefshort{suppsec:hyper} for more details). Then, the following loss is minimized for training:
 
  \vspace{-1.5em}
\begin{equation}
    \mathcal{L} = \frac{1}{N}\sum_{i=1}^{N}\|p(t) -  \Tilde{p}(t)\| + \lambda_1 TV(\mathcal{Z}(t)) +  \lambda_2 TV(\mathcal{C}(t)),
\label{eq:loss}
\end{equation}

where $p$ is the ground truth pixel color, and $TV(\mathcal{Z}(t))$ and  $TV(\mathcal{C}(t))$ are the total variation losses on the density and light fields. $\lambda_1, \lambda_2$ are hyperparameters.

Two important remarks are in order: \textit{a}) our model only requires the TV loss as a  loss regularizer, as opposed to multiple explicit regularizations that are used in many existing dynamic NeRF architectures such as explicit foreground-background modeling\cite{tretschk2021non, gao2021dynamic}, energy-preservation \cite{park2021nerfies}, or temporal consistency losses \cite{li2021neural, wang2021neural}. \textit{b}) To address the insufficiency of neural priors in regularizing the architecture, many dynamic NeRF methods tend to adopt cumbersome training procedures to converge to a good minimum, \textit{e.g.}, sequential training of temporally-ordered frames \cite{pumarola2021d, li2021neural}, coarse-to-fine annealing of hyperparameters \cite{park2021nerfies, park2021hypernerf}, or morphology processing \cite{yoon2020novel}. In contrast, we simply randomly sample points in time and space and feed them to the model for training. We argue that this is a strong indicator of the well-built inductive bias/implicit regularization of our architecture and the stability of our formulation.

 \vspace{-0.2em}
\section{Experiments}
\label{sec:experiments}
\vspace{-0.1em}

\begin{figure*}[!htp]
\centering
\includegraphics[width=0.7\textwidth]{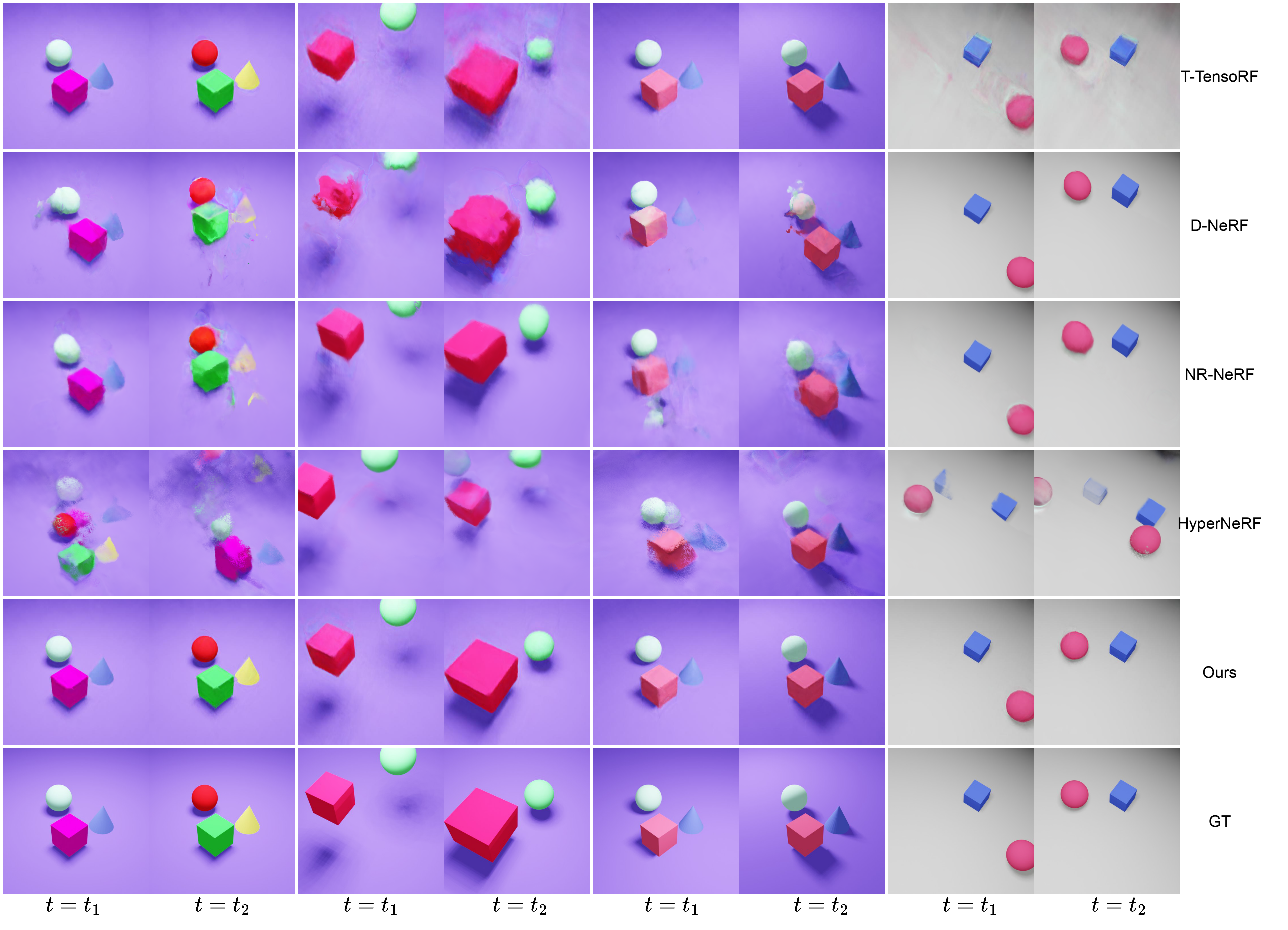}
 \vspace{-1em}
\caption{\small \textbf{Qualitative results on the synthetic dataset.} A comparison of novel views rendered at unseen times. As is evident, D-NeRF, NR-NeRF and HyperNeRF fail to accurately infer the 3D structure of the scenes containing texture and lighting changes (columns $1,2,5,6$). This behavior is caused by their inability to precisely disentangle light and density fields (see Sec.~\ref{subsec:limitations}). In contrast, T-NeRF performs relatively well in these scenes as it  achieves this disentanglement by construction. However, all the baselines exhibit poor reconstructions in the scale change and ball move scenes (columns $3,4,7,8$). This is an illustration of the sub-optimal localization of motion caused by inferior factorization of time and space, that are built into these models. Further, note that the objects in all the scenes are slightly misaligned in the baseline reconstructions, demonstrating sub-par disentanglement between scene and camera dynamics. The proposed model yields significantly higher quality new views.}
\label{fig:synthetic}
\end{figure*}

\begin{table*}[htbp]
    \centering
    \footnotesize
    \ra{1}
    \renewcommand\tabcolsep{2.5pt}
    \begin{tabularx}{0.85\textwidth}{lcccccccccccccccc}
        \toprule
        && \multicolumn{3}{c}{Cat} && \multicolumn{3}{c}{Climbing} && \multicolumn{3}{c}{Flashlight} && \multicolumn{3}{c}{Flower} \\
        Method && PSNR$\uparrow$ & SSIM$\uparrow$ & LPIPS$\downarrow$ && PSNR$\uparrow$ & SSIM$\uparrow$ & LPIPS$\downarrow$ && PSNR$\uparrow$ & SSIM$\uparrow$ & LPIPS$\downarrow$ && PSNR$\uparrow$ & SSIM$\uparrow$ & LPIPS$\downarrow$  \\
        \cmidrule{1-1} \cmidrule{3-5} \cmidrule{7-9} \cmidrule{11-13} \cmidrule{15-17}
        TensoRF && 24.05 & 0.81 & 0.35 && 22.53 & 0.76 & 0.34 && 26.70 & 0.90 & 0.36 && 26.36 & 0.86 & 0.29  \\
        T-TensoRF && 29.45 & 0.88 & 0.24 && 27.08 & 0.81 & 0.30 && 28.93 & 0.91 & 0.31 && 27.10 & 0.85 & 0.33  \\
        D-NeRF && 27.49 & 0.86 & 0.30 && 28.90 & 0.85 & 0.27 && 31.79 & \textbf{0.95} & 0.23 && 28.56 & \textbf{0.90} & 0.25  \\
        NR-NeRF && 26.63 & 0.82 & 0.35 && 25.59 & 0.79 & 0.35 && 30.59 & 0.93 & 0.29 && 25.57 & 0.84 & 0.37  \\
        HyperNeRF && 17.13 & 0.64 & 0.52 && 17.66 & 0.71 & 0.41 && 23.51 & 0.89 & 0.41 && 22.74 & 0.83 & 0.36  \\
        TiNeuVox && 22.41 & 0.81 & 0.36 && 24.99 & 0.81 & 0.35 && 31.13 & 0.91 & 0.27 && 28.11 & 0.88 & 0.27\\
        \ours (ours) && \textbf{29.69} & \textbf{0.89} & \textbf{0.21} && \textbf{29.14} & \textbf{0.86} & \textbf{0.25} && \textbf{31.80} & \textbf{0.95} & \textbf{0.19} && \textbf{29.98} & \textbf{0.90} & \textbf{0.22}  \\
        \midrule
        && \multicolumn{3}{c}{Color Change} && \multicolumn{3}{c}{Falling and Scale} && \multicolumn{3}{c}{Light Move} && \multicolumn{3}{c}{Ball Move} \\
        Method && PSNR$\uparrow$ & SSIM$\uparrow$ & LPIPS$\downarrow$ && PSNR$\uparrow$ & SSIM$\uparrow$ & LPIPS$\downarrow$ && PSNR$\uparrow$ & SSIM$\uparrow$ & LPIPS$\downarrow$ && PSNR$\uparrow$ & SSIM$\uparrow$ & LPIPS$\downarrow$  \\
        \cmidrule{1-1} \cmidrule{3-5} \cmidrule{7-9} \cmidrule{11-13} \cmidrule{15-17}
        TensoRF && 19.08 & 0.89 & 0.22 && 17.30 & 0.86 & 0.35 && 19.61 & 0.79 & 0.47 && 24.31 & 0.94 & 0.28  \\
        T-TensoRF && 35.16 & \textbf{0.97} & 0.09 && 24.30 & 0.90 & 0.32 && 36.49 & 0.97 & \textbf{0.10} && 28.27 & 0.96 & 0.33  \\
        D-NeRF && 17.42 & 0.89 & 0.28 && 24.60 & 0.92 & 0.23 && 19.15 & 0.91 & 0.25 && 22.58 & 0.95 & 0.20  \\
        NR-NeRF && 16.37 & 0.89 & 0.27 && 15.97 & 0.86 & 0.26 && 18.55 & 0.91 & 0.26 && 23.21 & 0.95 & 0.21  \\
        HyperNeRF && 16.19 & 0.84 & 0.35 && 14.46 & 0.83 & 0.34 && 16.10 & 0.83 & 0.40 && 20.27 & 0.93 & 0.27  \\
        TiNeuVox && 17.01 & 0.84 & 0.31 && 16.19 & 0.86 & 0.22 && 15.01 & 0.81 & 0.38 && 22.41 & 0.95 & 0.23\\
        \ours (ours) && \textbf{36.68} & \textbf{0.97} & \textbf{0.08} && \textbf{35.74} & \textbf{0.97} & \textbf{0.11} && \textbf{38.04} & \textbf{0.98} & \textbf{0.10} && \textbf{39.32} & \textbf{0.99} & \textbf{0.09}  \\

        \bottomrule
    \end{tabularx}
     \vspace{-1em}
    \caption{\small Quantitative comparison of novel view synthesis on our real and synthetic datasets. We report the average PSNR, SSIM (higher is better) and LPIPS (lower is better) results. Best results are in bold.}
    \label{tab:quant-our}
\end{table*}
 \vspace{-1em} 



\noindent \textbf{Datasets: } We collect four synthetic scenes and four real-world scenes as our dataset. The synthetic scenes include texture changes, lighting changes, scale changes, and long-range movements. The real-world scenes include lighting changes, long-range movements, and spatially concentrated dynamic objects (\supprefshort{suppsec:datasets}). To demonstrate the ability of our method to capture topologically varying deformations, we also evaluate against the HyperNeRF dataset \cite{park2021hypernerf}.

\noindent \textbf{Baselines:} We choose D-NeRF~\cite{pumarola2021d}, NR-NeRF~\cite{tretschk2021non}, TiNeuVox \cite{TiNeuVox} and HyperNeRF \cite{park2021hypernerf} as our main baselines. All are recently proposed Dynamic-NeRF models that adopt the ray deformation paradigm. NR-NeRF  comprises an explicit neural network for isolating the motion of a scene, and HyperNeRF consists of separate MLPs for modeling time and space deformations, providing ideal baselines for evaluating the efficacy of our space-time priors. For baselines, we performed a grid search for the optimal hyperparameters for each scene. In contrast, our model uses a \textbf{single hyperparameter setting} across all the scenes, demonstrating its robustness (see \supprefshort{suppsec:hyper} for hyperparameter and training details). Further, it is essential to  validate whether the superior performance of our model stems from the light/density disentanglement or the space-time factorization. Thus, we design another baseline T-TensoRF, which disentangles the light and density fields, but do not factorize time and space dynamics (see \supprefshort{suppsec:tnerf}).

\begin{figure*}[!htp]
\centering
\includegraphics[width=.9\textwidth]{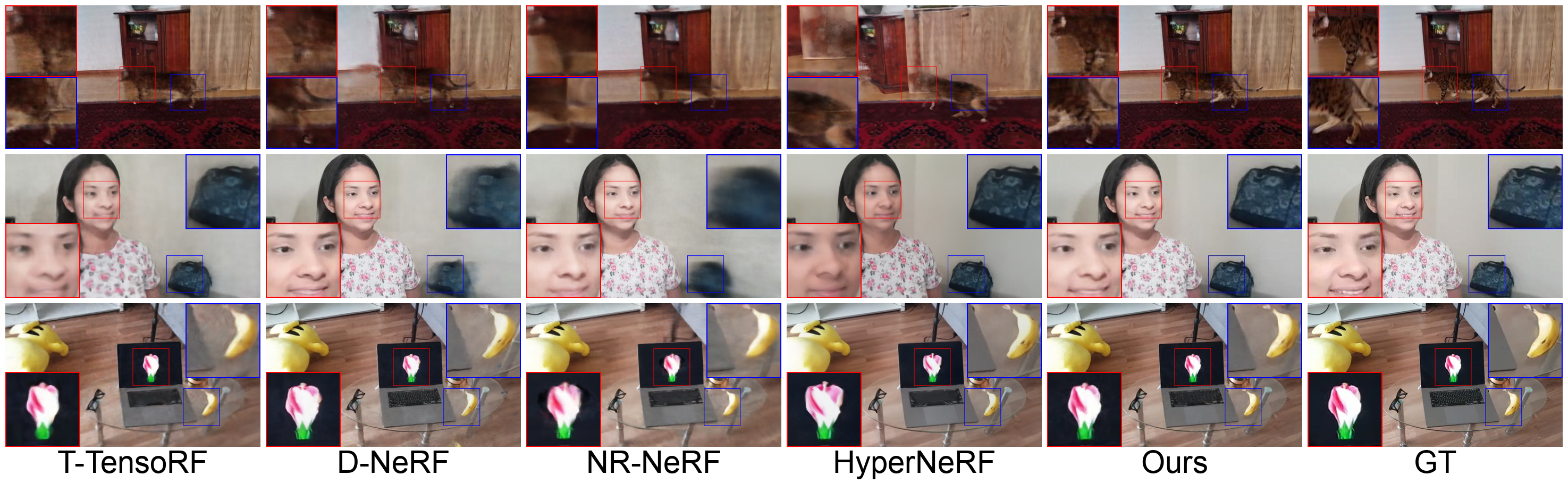}
 \vspace{-1em}
\caption{\small \textbf{Qualitative comparison on the real-world dataset (zoom-in for a better view).} The shown examples represent novel views generated at unseen time instances. Note that in the flashlight scene (second row), D-NeRF, NR-NeRF and T-NeRF fail to capture high-fidelity details in the background. On the other hand, in the cat-walking scene (top row) where the object moves across a considerable range in space, all the baselines fail to recover the moving object accurately. In the flower scene (third row), where the motion is constrained within a small region, the baselines perform fairly well. In comparison, our method exhibits superior performance in all the cases.}
\label{fig:real}
\end{figure*}

\begin{table*}[htbp]
    \centering
    \footnotesize
    \ra{1}
    \renewcommand\tabcolsep{3pt}
    \begin{tabularx}{0.85\textwidth}{lcccccccccccccccc}
        \toprule
        && \multicolumn{3}{c}{Color Change} && \multicolumn{3}{c}{Falling and Scale} && \multicolumn{3}{c}{Light Move} && \multicolumn{3}{c}{Ball Move} \\
        Basis && PSNR$\uparrow$ & SSIM$\uparrow$ & LPIPS$\downarrow$ && PSNR$\uparrow$ & SSIM$\uparrow$ & LPIPS$\downarrow$ && PSNR$\uparrow$ & SSIM$\uparrow$ & LPIPS$\downarrow$ && PSNR$\uparrow$ & SSIM$\uparrow$ & LPIPS$\downarrow$  \\
        \cmidrule{1-1} \cmidrule{3-5} \cmidrule{7-9} \cmidrule{11-13} \cmidrule{15-17}
        DCT  && 33.99 & 0.93 & 0.14 && 32.61 & 0.89 & 0.19 && 33.77 & 0.92 & 0.16 && 33.59 & 0.93 & 0.16  \\
        Fourier && 31.33 & 0.89 & 0.19 && 29.74 & 0.89 & 0.21 && 31.99 & 0.91 & 0.23 && 33.45 & 0.94 & 0.19 \\
        Bernstein && 27.81 & 0.86 & 0.21 && 28.90 & 0.87 & 0.25 && 31.57 & 0.91 & 0.25 && 33.53 & 0.94 & 0.18  \\
         Neural && \textbf{36.68} & \textbf{0.97} & \textbf{0.08} && \textbf{35.74} & \textbf{0.97} & \textbf{0.11} && \textbf{38.04} & \textbf{0.98} & \textbf{0.10} && \textbf{39.32} & \textbf{0.99} & \textbf{0.09}  \\

        \bottomrule
    \end{tabularx}
    \vspace{-1em}
    \caption{\small \textbf{Ablation of different time-basis functions.} Although other basis functions are able to yield acceptable performances, learned neural basis functions perform better.}
    \label{tab:abl-basis}
\end{table*}
\vspace{-1em}

\begin{table*}[!htbp]
    \centering
    \footnotesize
    \ra{1}
    \renewcommand\tabcolsep{3pt}
    \begin{tabularx}{0.95\textwidth}{lccccccccccccccccccc}
        \toprule
        && \multicolumn{2}{c}{Expressions} && \multicolumn{2}{c}{Teapot} && \multicolumn{2}{c}{Chicken} && \multicolumn{2}{c}{Fist} && \multicolumn{2}{c}{Banana} && \multicolumn{2}{c}{Lemon} \\
         && PSNR$\uparrow$  & LPIPS$\downarrow$ && PSNR$\uparrow$ & LPIPS$\downarrow$ && PSNR$\uparrow$  & LPIPS$\downarrow$ && PSNR$\uparrow$  & LPIPS$\downarrow$ && PSNR$\uparrow$  & LPIPS$\downarrow$ && PSNR$\uparrow$  & LPIPS$\downarrow$ \\
        \cmidrule{1-1} \cmidrule{3-4} \cmidrule{6-7} \cmidrule{9-10} \cmidrule{12-13}
        \cmidrule{15-16}
        \cmidrule{18-19}
       NV \cite{lombardi2019neural} && 26.7 & 0.215 && 26.2 & 0.216 && 22.6 & 0.243 && 29.3 & 0.213 && 24.8 & 0.209 && 28.8 & 0.190 \\
       NSFF \cite{li2021neural} && 26.6 & 0.283 && 25.8 & \textbf{0.210} && 27.7 & 0.173 && 24.9 & 0.329 && 26.1 & 0.243 && 28.0 & 0.283 \\
       Nerfies \cite{park2021nerfies} && 27.5 & 0.224 && 25.7 & 0.225 && 28.7 & 0.141 && 29.9 & 0.171 && 27.9 & 0.209 && 30.8 & 0.223 \\
       HyperNeRF \cite{park2021hypernerf} && 27.9 & 0.218 && \textbf{26.4} & 0.212 && 28.7 & 0.156 && \textbf{30.7} & \textbf{0.150} && \textbf{28.4} & 0.191 && 31.8 & \textbf{0.210 }\\
       \ours (ours) && \textbf{28.2} & \textbf{0.213} && 26.1 & 0.215 && \textbf{29.8} & \textbf{0.141} && 28.4 & 0.161 && 28.2 & \textbf{0.191} && \textbf{32.1} & 0.223\\

        \bottomrule
    \end{tabularx}
    \vspace{-1em}
    \caption{\small \textbf{Comparison on the HyperNeRF dataset.} Numbers for the competing methods are extracted from \cite{park2021hypernerf}.}
    \label{tab:hyper}
\end{table*}

\subsection{Synthetic scenes } 
\vspace{-0.5em}
The synthetic scenes consist of four scenes: \emph{texture change, falling and scale, light move, and ball move}. See Fig.~\ref{fig:synthetic} for a qualitative comparison. As shown, D-NeRF, NR-NeRF, and HyperNeRF fail to accurately model the color and light changes. This validates our claim in Sec.~\ref{sec:ray_bending}, that for full disentanglement of light and density fields, the above methods require a block diagonal Jacobian structure, which is an extremely restrictive condition. Similarly, they tend to deform the objects when scale changes and long-range movements are present. T-TensoRF, due its ability to disentangle light and density fields, adequately recovers light/texture changes. However, all the baselines fail to accurately learn the 3D positions of the objects showcasing their inability to precisely disentangle camera and scene dynamics.  In comparison, our method achieves significantly superior results in all above aspects. See Table~\ref{tab:quant-our} for quantitative results.

 \vspace{-0.2em}
\subsection{Real-world scenes }
\vspace{-0.5em}
The real-world scenes contain four scenes; \emph{cat walking, flashlight, flower, and  climbing}. Cat walking and climbing scenes contain long-range movements. See Fig.~\ref{fig:real} and Fig.~\ref{fig:teaser} for qualitative comparisons on these scenes. When long-range movements are present, the baselines fail to recover the high-fidelity details of the moving objects. In the flashlight scene, baselines fail to accurately capture granular details in the background or light change. In the flower scene, where the dynamics are concentrated spatially, the baselines perform  well. Our method generates better results in all of the aforementioned aspects. Note that D-NeRF, NR-NeRF can model lighting changes as shown in the flashlight scene (see also \supprefshort{suppsec:comparisons}), validating our insights in Sec.~\ref{sec:ray_bending} that ray deformation models indeed encode  density and light field dynamics. Table~\ref{tab:quant-our} depicts quantitative results. 

\subsection{Topologically varying scenes}
\vspace{-0.5em}
Park \etal \cite{park2021hypernerf} showed that most existing dynamic NeRF methods cannot model topologically varying scenes effectively. To remedy this, they proposed a method that models discontinuities of the evolving field as continuous deformations, using a collection of MLPs. In contrast, our method can implicitly model such scenes since we model the evolution of each 3D point in the fields as bandlimited signals. To showcase this, we conduct experiments on the topologically varying interpolation dataset provided by \cite{park2021hypernerf}. 
In contrast to our dataset, these scenes exhibit a camera revolving around quasi-repeated actions which adhere to the camera motion and object centric biases baked into ray-deformation methods (Sec. \ref{sec:ray_bending}), and do not include long-range motions or light/texture changes. The results are depicted in Table \ref{tab:hyper}. As is shown, we achieve near-identical\footnote{HyperNeRF\cite{park2021hypernerf} uses distortion coefficients to correct the rays, we omit this detail from our implementation to maintain a fair comparison with the other baselines.} or better results compared to baselines.

\subsection{Ablation study}
\vspace{-0.75em}
Our generic framework allows different implementations. Thus, we compare other possible time-basis functions that are complete in $L^2(\mathbb{R}, dt)$ against the neural trajectory basis. Table~\ref{tab:abl-basis} presents a quantitative comparison with the DCT, Fourier, and Bernstein bases. Although these basis functions are also capable of providing acceptable results, neural basis performs best.  We  provide ablations for other design choices as well; manifold regularization, \# basis functions, neural prior, and low rank factorization in \supprefshort{suppsec:ablations}. T-TensoRF demonstrates the effect of factorization of space-time dynamics. 
\vspace{-3pt}


\vspace{-0.5em}
\section{Limitations}
\label{sec:limitations}
\vspace{-0.5em}

As we use a volumetric representation instead of an MLP to model the fields, resolution of the reconstruction is limited by that of the volumetric representation. Increasing the resolution leads to higher memory consumption, which is a common attribute of all grid-based NeRF models \cite{yu2021plenoxels, chen2022tensorf, muller2022instant} that sacrifice memory for speed. 
\vspace{-1em}
\section{Conclusion}
\label{sec:conclusions}
\vspace{-0.5em}
We offer a novel, generic framework for modeling dynamic 3D scenes which allows efficient factorization of the space and time dynamics. This factorization presents a platform to impose well-designed space-time priors (inspired by \nrsfm) on NeRF, enabling high-fidelity novel view synthesis of dynamics scenes. Finally, we present an implementation of the proposed framework that demonstrates compelling results across complex dynamics scenes containing long-range movements, scale changes, and light/texture changes.



{\small
\bibliographystyle{ieee_fullname}
\bibliography{main}
}

\clearpage
\newpage

\onecolumn

\begin{center}
    \section*{\Huge Supplementary Materials}
\end{center}

\addcontentsline{toc}{section}{Supplementary}
\renewcommand{\thesection}{\Alph{section}}

\setcounter{section}{0}

\section{Ray Deformation Networks}
\label{supsec:ray_bending}
\label{suppsec:ray_deformation}
\subsection{Ray deformation networks learn field dynamics}

In this section, we show evidence that the design methodology adopted by existing works for implementing the ray deformation framework do not learn point trajectories in space, and instead, act as light and density deformation modules. Consider an MLP $\Psi^d:(x,y,z,t) \to (\Delta x, \Delta y, \Delta z)$ outputting the transformation of a point $(x,y,z)$   at time instant $t$ with respect to a canonical setting. Then, a second MLP $\Psi^c:(x+\Delta x, y+\Delta y, z+\Delta z, \textbf{d}) \to (c, \sigma)$ takes in the deformed inputs and the viewing direction $\textbf{d}$, and predicts the density and light fields $(c, \sigma)$. However, one can interpret the above pipeline from another perspective. Observe that $\Psi^d$ and $\Psi^c$ can be considered as a single deep MLP $\Psi^{d \wedge c}: (x,y,z,t) \to (c, \sigma)$, where the bottleneck is three-dimensional. Further, there exists a skip connection from $(x,y,z)$ to the bottleneck. From this perspective, the above implementation is simply an MLP with a skip connection and bottleneck of dimension three, modeling a function from $(x,y,z,t)$ to $(c, \sigma)$. See Fig.~\ref{fig:ray_deform_network} for a visual illustration of this interpretation.  We empirically solidify this argument by showing that such networks can indeed model light and density deformations individually (to an extent), which is impossible with a model that only learns point movements in space,  according to the ray deformation framework (see Fig.~\ref{fig:lightdensity}). Next, we discuss limitations of ray deformation networks.

\begin{figure}[!htp]
\centering
\includegraphics[width=1.0\columnwidth]{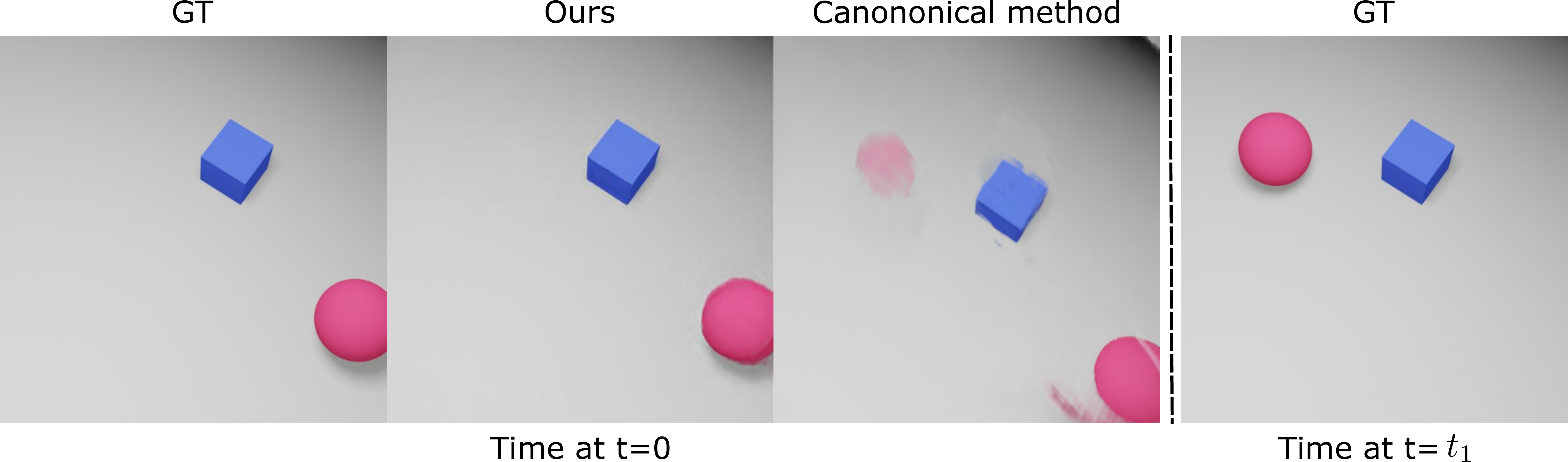}
\caption{\textbf{Our method does not rely on a canonical scene configuration}. Existing ray deformation models require choosing a canonical scene configuration at a user-defined time instance, which can hinder their performance in scenes where new information appears in subsequent frames. In contrast, our model does not suffer from such a limitation. }
\label{fig:canon_fail}
\vspace{-10pt}
\end{figure}

\begin{figure}[!htp]
\centering
\includegraphics[width=1.0\columnwidth]{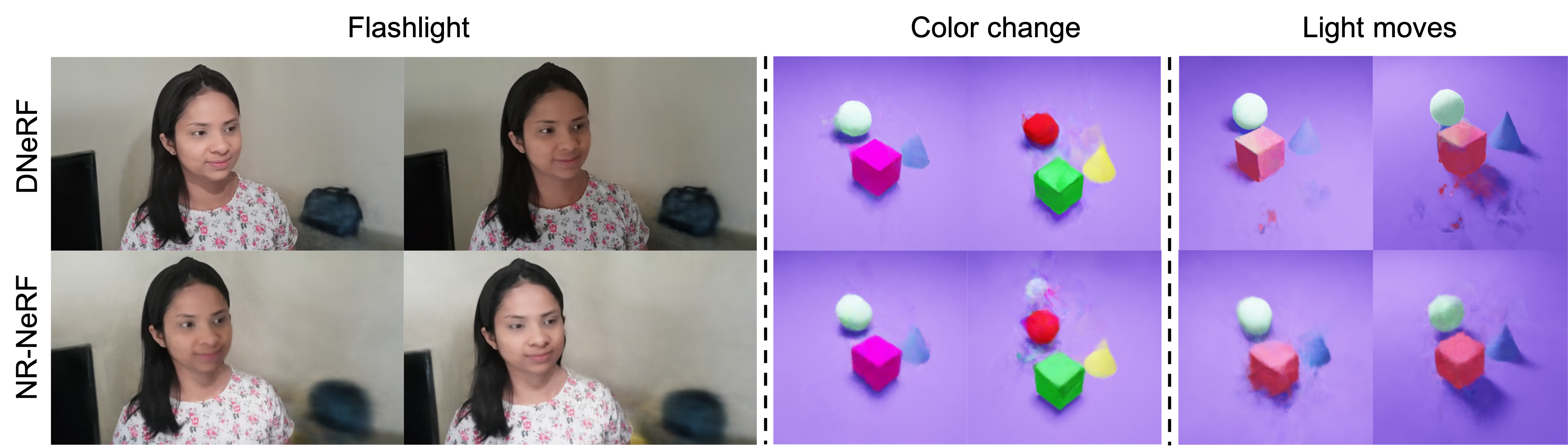}
\caption{\textbf{Ray deformation networks parameterize light and density fields  instead of ray deformations.} We show evidence for this using three example scenes. From left to right,  \textit{1)} A real world scene with light changes on the person's face. \textit{2)} The colors of the shapes are changing. \textit{3)} The light is shifting position in the scene. We can observe both ray deformation works, DNeRF\cite{pumarola2021d} and NR-NeRF\cite{tretschk2021non}, learn the texture changes that occur to an extent, which is not possible by simply learning ray deformations. However, the reconstructions are still sub-par as light and  density dynamics are entangled in these frameworks.}
\label{fig:lightdensity}
\vspace{-10pt}
\end{figure}

\begin{figure}[!htp]
\centering
\includegraphics[width=1.0\columnwidth]{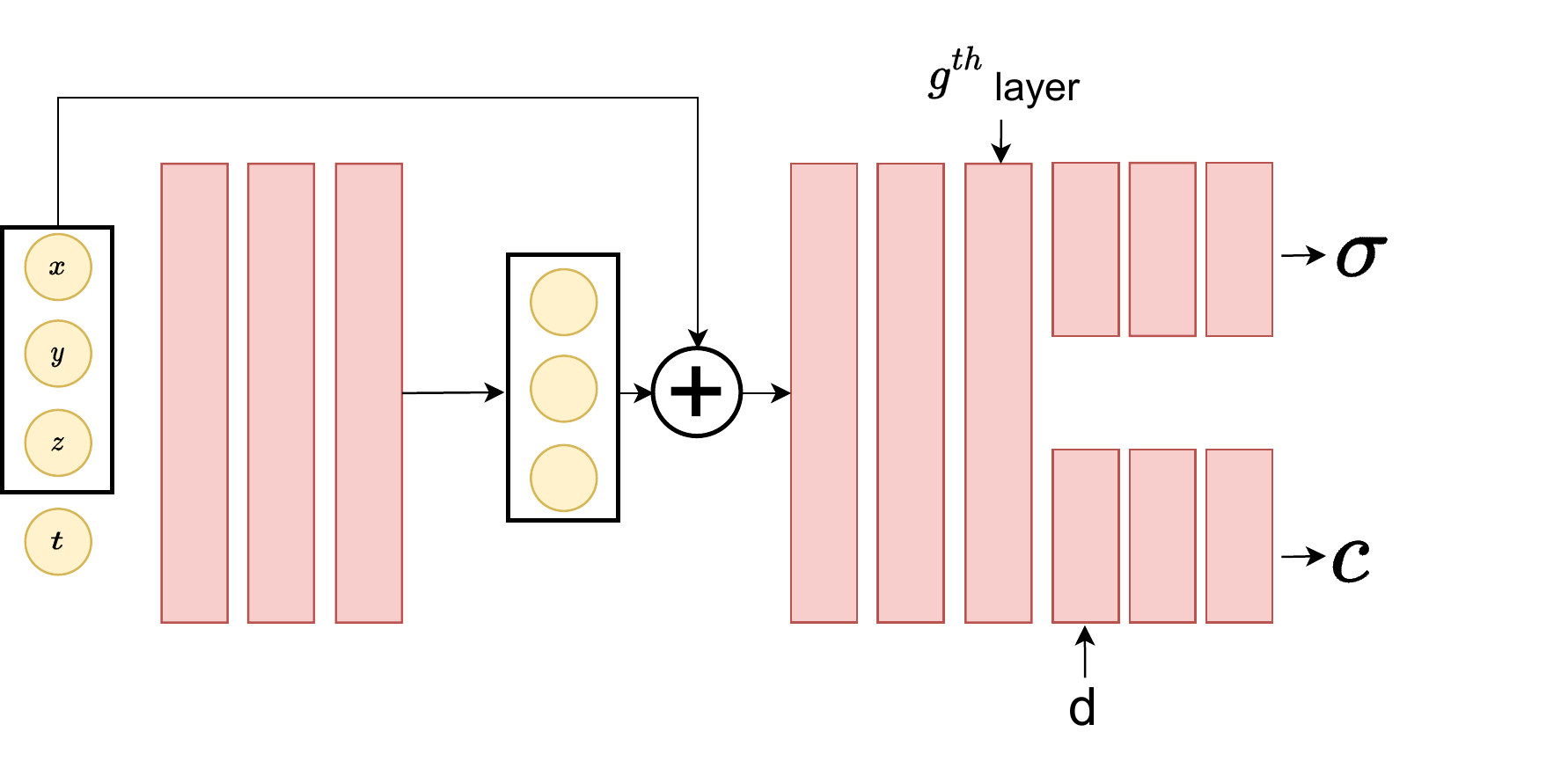}
\caption{\textbf{Ray deformation networks can be interpreted as a single deep network with a three-dimensional bottleneck.} With this interpretation, it is clear that ray deformation networks can indeed learn light and density evolution independently (to an extent), instead of simply learning ray deformations.}
\label{fig:ray_deform_network}
\vspace{-10pt}
\end{figure}

\subsection{Dependency on a canonical frame}

A weakness that comes with learning a canonical frame is one root cause for ray deformation networks to struggle in scenes with an object that has long translations. This type of formulation is also working against the smoothness assumptions of neural networks, which are now required to learn non-smooth representations. In Fig.~\ref{fig:canon_fail} we can see a toy example of a ball moving in a constant trajectory across the scene. The canonical method at time step $t=0$ is forced to learn an average representation of the scene and is unable to correctly represent the canonical frame which corresponds to image $GT$ at time $t = 0$. This formulation also negatively impacts ray deformation models to encode fine detail information, due to the constant averaging the canonical frame maintains throughout time.
In Fig.~1 (main paper) it can be seen that NR-NeRF\cite{tretschk2021non} fails to capture fine details such as shirt wrinkles and the climbers legs.

\subsection{Entanglement of light and density fields}

Let the output of the $g^{th}$ layer of Fig.~\ref{fig:ray_deform_network} be $g(\mathbf{x}):\mathbb{R}^4 \to \mathbb{R}^C$. Further, let $\psi_l:\mathbb{R}^C \to \mathbb{R}$ and  $\psi_d:\mathbb{R}^C \to \mathbb{R}$ be network branches that predict light and density fields, respectively, taking $g(\mathbf{x})$ as input. Now, consider a scenario where the light of the scene or the texture of objects change, while the objects remain static. In this case, we need the light field to be a function of time, while the density field should remain constant. Consider the Jacobians,

\begin{equation}
\mathbf{J}_{g} = 
\begin{bmatrix}
  \frac{\partial g_1(\mathbf{x})}{\partial x} &  \frac{\partial g_1(\mathbf{x})}{\partial y} & \frac{\partial g_1(\mathbf{x})}{\partial z} & \frac{\partial g_1(\mathbf{x})}{\partial t} \\
  \vdots & \vdots & \vdots & \vdots \\
  \frac{\partial g_C(\mathbf{x})}{\partial x} &  \frac{\partial g_C(\mathbf{x})}{\partial y} & \frac{\partial g_C(\mathbf{x})}{\partial z} & \frac{\partial g_C(\mathbf{x})}{\partial t}
\end{bmatrix},
\end{equation}

\begin{equation}
\mathbf{J}_{\psi_d} = 
    \begin{bmatrix}
  \frac{\partial \psi_d(g({x}))}{\partial g_1(\mathbf{x})} &  \frac{\partial \psi_d(g({x}))}{\partial g_2(\mathbf{x})} & \dots  \frac{\partial \psi_d(g({x}))}{\partial g_C(\mathbf{x})}. 
\end{bmatrix}
\end{equation}

Then, the Jacobian of $\psi_d \circ g$ becomes,
\begin{equation}
\label{equ:jac}
\mathbf{J}_{\psi_d \circ g} = 
    \begin{bmatrix}
  \frac{\partial \psi_d(g(\mathbf{x}))}{\partial x} &  \frac{\partial \psi_d(g(\mathbf{x}))}{\partial y} &  \frac{\partial \psi_d(g(\mathbf{x}))}{\partial z}  & \frac{\partial \psi_d(g(\mathbf{x}))}{\partial t}
\end{bmatrix} = \mathbf{J}_{\psi_d} \mathbf{J}_{g}.
\end{equation}

And, we need $\frac{\partial \psi_d(g({x}))}{\partial t} = 0$ since the density is not a function of time. Therefore, the $4^{th}$ column of $\mathbf{J}_g$ has to be orthogonal to $\mathbf{J}_{\psi_d}$. This can be achieved via one of the following three scenarios:
\begin{itemize}
    \item \textbf{Scenario 1:} \textit{$\mathbf{J}_{\psi_d}$ is a zero vector. }
    \item \textbf{Scenario 2:} \textit{The $4^{th}$ column of $\mathbf{J}_g$ is zero.}
    \item \textbf{Scenario 3:} \textit{Both scenario 1 and 2 are false, but $\mathbf{J}_{\psi_d}$ is orthogonal to the $4^{th}$ column of $\mathbf{J}_g$.}
\end{itemize}

However, Scenario 1 implies that $\frac{\partial \psi_d \circ g(\textbf{x})}{\partial x,y,z}$ is zero (see Eq.~\ref{equ:jac}), which makes the density constant across space. On the other hand, Scenario 2 implies that $\frac{\partial \psi_l}{\partial t}$ is zero since $\frac{\partial \psi_l}{
\partial t} = \frac{\partial \psi_l}{\partial g}\frac{\partial g}{\partial t}$. That is, with Scenario 2, the light cannot be a function of time. Further, in general, Scenario 3 makes $\mathbf{J}_{\psi_d}$ a function of  $\frac{\partial {g(\mathbf{x})}}{\partial t}$ since in the case where $g$ obeys the following PDE,

\begin{equation}
\frac{\partial {g(\mathbf{x})}}{\partial t} = q(t),
\end{equation}

where $q(t)$ is some function parameterized by $t$. Thus, it is clear that $\mathbf{J}_{\psi_d}$ becomes a function of $t$. On the other hand, by Eq.~\ref{equ:jac}, $\frac{\partial \psi_d \circ g(\textbf{x})}{\partial x,y,z}$ also becomes a function of time, unless both $\mathbf{J}_{\psi_d}$ and $\mathbf{J}_g$ preserves a block structure such that

\begin{equation}
\mathbf{J}_{g} = 
\begin{bmatrix}
  \frac{\partial g_1(\mathbf{x})}{\partial x} &  \frac{\partial g_1(\mathbf{x})}{\partial y} & \frac{\partial g_1(\mathbf{x})}{\partial z} & 0\\
  \vdots & \vdots & \vdots & \vdots \\
  \frac{\partial g_c(\mathbf{x})}{\partial x} &  \frac{\partial g_c(\mathbf{x})}{\partial y} & \frac{\partial g_c(\mathbf{x})}{\partial z} & 0 \\
  0 & 0 & 0 & \frac{\partial g_{c+1}(\mathbf{x})}{\partial t}  \\
  \vdots & \vdots & \vdots & \vdots \\
  0 & 0 & 0 &  \frac{\partial g_C(\mathbf{x})}{\partial t}
\end{bmatrix},
\end{equation}

and 

\begin{equation}
\mathbf{J}_{\psi_d} = 
    \begin{bmatrix}
  \frac{\partial \psi_d(g{x})}{\partial g_1(\mathbf{x})}  & \dots  \frac{\partial \psi_d(g{x})}{\partial g_c(\mathbf{x})} & 0 & \dots & 0
\end{bmatrix}
\end{equation}

Note that this is an extremely unique solution that is seldom achieved in practice under general conditions, due to the ill-posed nature of the problem. In most cases, the networks tend to converge to solutions where the $4^{th}$ column of $\mathbf{J}_g$ becomes non-zero, in order to model the light changes, which in turn makes the density a function of time. This causes an inherent entanglement of the light and density fields. The toy example results shown in Fig.~\ref{fig:lightdensity} is an illustration of this behavior.

\subsection{Limited expressiveness}

Consider the parameterization of the density field. As evident from Fig.~\ref{fig:ray_deform_network}, it is modeled with a network with a bottleneck of dimension three. In this setting, the density field becomes a manifold of dimension three. In other words, the dynamics of the density field can be modeled with only three parameters. However, recall that in complex scenes, particular points of the density field may need to be parameterized by $(x,y,z,t)$ simultaneously. In other words, it is required that $\frac{\partial \sigma}{\partial x}, \frac{\partial \sigma}{\partial y}, \frac{\partial \sigma}{\partial z}, \frac{\partial \sigma}{\partial t} \neq 0$ at some points in space-time. Thus, having a bottleneck of three hinders the network from modeling such complex dynamics.

\subsection{Entanglement of space and temporal variations}

A ray deformation network can be considered as a single deep network with bottleneck three, that takes $(x,y,z,t)$ as inputs and models light/density field deformations. However, often, space and time consist of contrasting spectral properties; objects deform smoothly across time, but space may contain sharp/high frequency variations. Therefore, using a single neural network to model this two extremes can be sub-optimal. Generally, a network with a higher bandwidth is ideal for modeling space, and a lower bandwidth is necessary for modeling time. 

\begin{figure}[!htp]
\centering
\includegraphics[width=1.0\columnwidth]{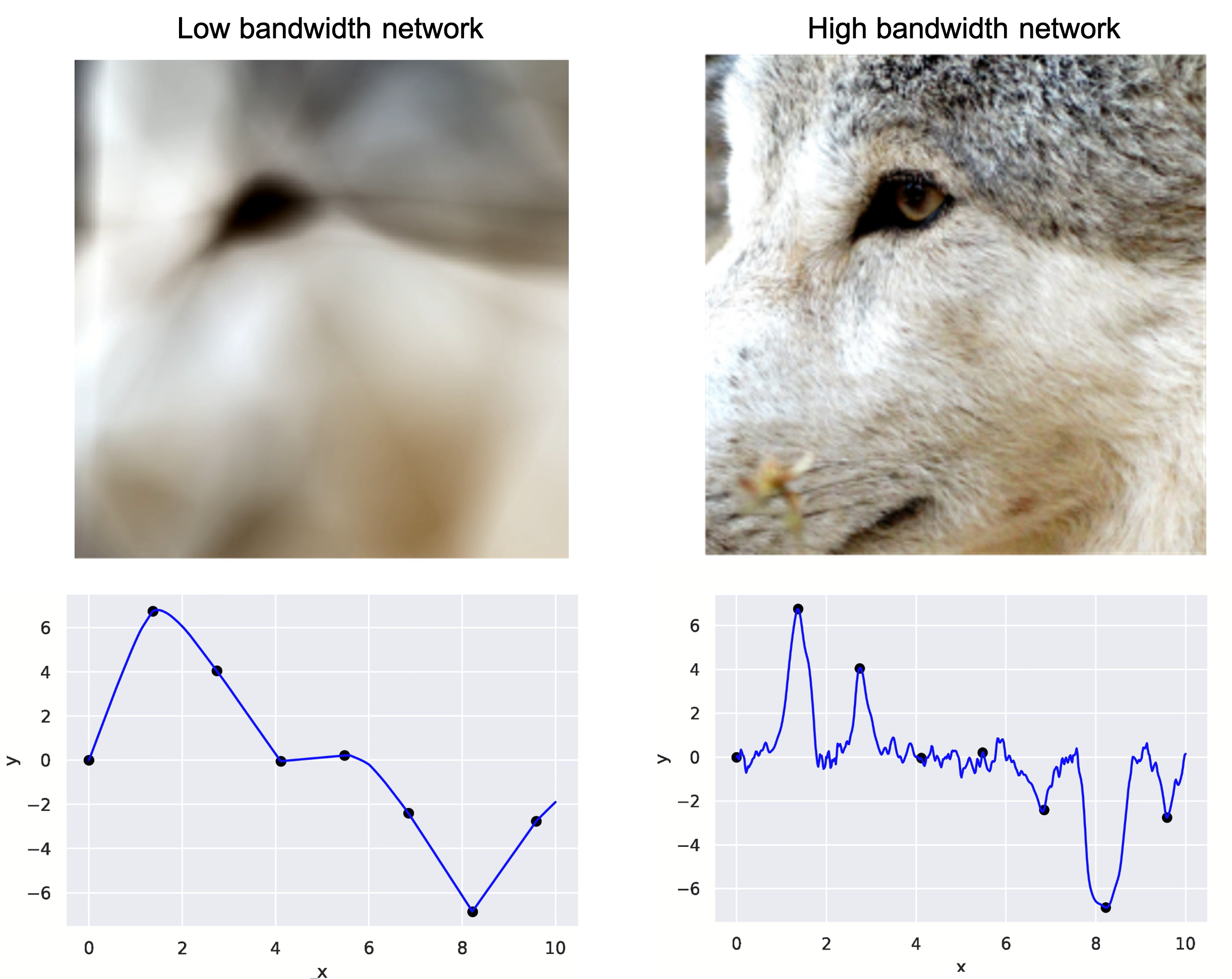}
\caption{\textit{Left column: } A low-bandwidth network cannot capture high-frequency content adequately,  which can be sub-optimal for modeling space. However, a low-bandwidth network is ideal for interpolating sparse low frequency points, which is optimal for modeling temporal dynamics. \textit{Right column: } A high-bandwidth network can reconstruct sharp variations, but results in erratic interpolations. This can be detrimental for smooth temporal dynamics modeling. We used four layer ReLU networks  with positional embeddings for encoding the signals. We obtained networks with different bandwidths by changing the frequency support of the positional embedding layer.}
\label{fig:entanglement}
\vspace{-10pt}
\end{figure}

Fig.~\ref{fig:entanglement} shows an illustration. Using a high bandwidth network for interpolating across time  allows a network to perfectly memorize training data, but can result in erratic interpolations. On the other hand, using a low-bandwidth network leads to low fidelity space reconstructions. Note that since space is typically more densely sampled (compared to time axis) in the dynamic NeRF setting, a high-bandwidth network can recover both low and high frequencies in space, i.e., supervision is available more densely. This is a key factor that motivates space/time factorization, as in our framework. This behavior was also observed previously by \cite{ramasinghe2022beyond} and \cite{ramasinghe2022you}.

\section{Ablations}
\label{suppsec:ablations}

In this section, we show ablations on manifold regularization, number of basis functions, neural prior, and low rank factorization. As seen in Table.~\ref{tab:abl-basis-number}, the performance of the model saturates at around $24$ basis functions. Further, the effect of manifold regularization is quite significant  (Table.~\ref{tab:abl-manifold}). Table \ref{tab:abl-lowrank} depicts the advantage of the low-rank prior. For removing the low-rank prior, we remove the VM decomposition from the implementation.

\begin{table*}[!htbp]
    \centering
  
    \ra{1}
    \renewcommand\tabcolsep{3pt}
    \begin{tabularx}{0.7\textwidth}{lcccccccccccc}
    \toprule
        && \multicolumn{2}{c}{Color Change} && \multicolumn{2}{c}{Falling and Scale} && \multicolumn{2}{c}{Light Move} && \multicolumn{2}{c}{Ball Move} \\
        \# Basis && PSNR$\uparrow$ & SSIM$\uparrow$  && PSNR$\uparrow$ & SSIM$\uparrow$  && PSNR$\uparrow$ & SSIM$\uparrow$  && PSNR$\uparrow$ & SSIM$\uparrow$   \\
        \cmidrule{1-1} \cmidrule{3-4} \cmidrule{6-7} \cmidrule{9-10} \cmidrule{12-13}
        4  && 33.16 & 0.96 && 12.04 & 0.81  && 13.44 & 0.81  && 13.59 & 0.82  \\
        12 && 35.19 & 0.95  && 28.18 & 0.89 && 27.50 & 0.88  && 26.91 & 0.88  \\
        24 && 36.68 & 0.97  && 35.74 & 0.97  && 38.04 & 0.98 && 39.32 & 0.99   \\
         48 && 36.61 & 0.97 && 35.19 & 0.97  && 38.04 & 0.98  && 39.33 & 0.99  \\
         \bottomrule
    \end{tabularx}
    \caption{Performance against the number of time-basis functions.}
    \label{tab:abl-basis-number}
\end{table*}

\begin{table*}[!htbp]
    \centering
  
    \ra{1}
    \renewcommand\tabcolsep{3pt}
    \begin{tabularx}{0.85\textwidth}{lcccccccccccc}
    \toprule
        && \multicolumn{2}{c}{Color Change} && \multicolumn{2}{c}{Falling and Scale} && \multicolumn{2}{c}{Light Move} && \multicolumn{2}{c}{Ball Move} \\
        \# Regularization && PSNR$\uparrow$ & SSIM$\uparrow$  && PSNR$\uparrow$ & SSIM$\uparrow$  && PSNR$\uparrow$ & SSIM$\uparrow$  && PSNR$\uparrow$ & SSIM$\uparrow$   \\
        \cmidrule{1-1} \cmidrule{3-4} \cmidrule{6-7} \cmidrule{9-10} \cmidrule{12-13}
        W/O manifold regularization && 33.11 & 0.96  && 32.16 & 0.96 && 38.00 & 0.98  && 36.17 & 0.97  \\
        W/manifold regularization && 36.68 & 0.97  && 35.74 & 0.97  && 38.04 & 0.98 && 39.32 & 0.99   \\
         \bottomrule
    \end{tabularx}
    \caption{The effect of manifold regularization.}
    \label{tab:abl-manifold}
\end{table*}

\begin{table*}[!htbp]
    \centering
  
    \ra{1}
    \renewcommand\tabcolsep{3pt}
    \begin{tabularx}{0.85\textwidth}{lcccccccccccc}
    \toprule
        && \multicolumn{2}{c}{Color Change} && \multicolumn{2}{c}{Falling and Scale} && \multicolumn{2}{c}{Light Move} && \multicolumn{2}{c}{Ball Move} \\
        \# Regularization && PSNR$\uparrow$ & SSIM$\uparrow$  && PSNR$\uparrow$ & SSIM$\uparrow$  && PSNR$\uparrow$ & SSIM$\uparrow$  && PSNR$\uparrow$ & SSIM$\uparrow$   \\
        \cmidrule{1-1} \cmidrule{3-4} \cmidrule{6-7} \cmidrule{9-10} \cmidrule{12-13}
        W/O low rank factorization && 34.33 & 0.96  && 31.19 & 0.95 && 37.99 & 0.96  && 37.99 & 0.97  \\
        W/ low rank factorization && 36.68 & 0.97  && 35.74 & 0.97  && 38.04 & 0.98 && 39.32 & 0.99   \\
         \bottomrule
    \end{tabularx}
    \caption{The effect of low rank factorization.}
    \label{tab:abl-lowrank}
\end{table*}

\subsection{Implicit regularization of the neural trajectory basis}
\label{suppsec:neural_basis}

This sub section discusses the effect of neural trajectory prior on the model performance. 
Using a combination of trajectory basis to reconstruct the motion of a set of points is popular in \nrsfm \cite{akhter2008nonrigid}. This technique implicitly restricts the solution to a known low-dimensional subspace of smooth trajectories. One such popular trajectory basis is the DCT basis. A key advantage of this method compared to a shape basis is that an object-agnostic basis can be employed across multiple scenes. However, although the basis type is scene agnostic, the basis dimensionality depends on multiple factors such as scene dynamics, camera dynamics, and sequence length \cite{park20113d}. Thus, the dimensionality of the basis functions should be tuned per scene.

In an attempt to solve the above problem, \cite{zhu20113d} applied an $\ell_1$ norm penalty on the coefficients of the trajectory basis. In practice, a sparse-coding algorithm \cite{lee2006efficient} was used to achieve this. Although this strategy was effective, it ignores an important prior; for natural signals, the DCT basis tends to concentrate of the lower frequencies. 

An alternative and a more effective way of regularizing the trajectory basis has been minimizing the trajectory responses to high-pass filters. \cite{valmadre2012general} showed that such regularization is able to enforce local temporal constraints, rather than global constraints, which extends trivially to sequences of different
length. They particularly showed that this mechanism alleviate the need to tune the basis size. This approach also has a physical interpretation; minimizing the $\ell_2$ norm of the second-order derivative is equivalent to an assumption of constant mass subject to isotropic Gaussian distributed forces \cite{salzmann2011physically}. Similarly, minimizing the $\ell_2$ norm of the first-order derivative is equivalent to finding the solution with the least kinetic energy. 

\begin{figure}[!htp]
\centering
\includegraphics[width=1.0\columnwidth]{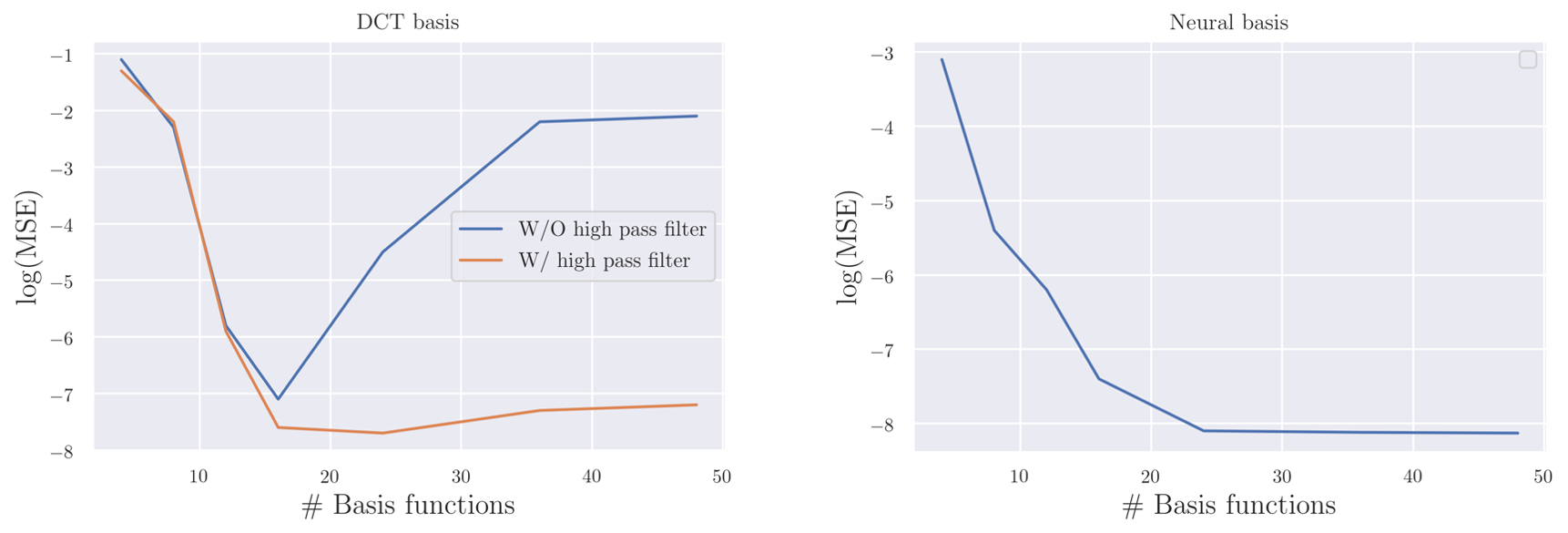}
\caption{\textbf{Implicit regularization of the neural trajectory basis.} \textit{Left:} With the DCT basis, the performance is sensitive to the number of basis functions. After an optimal basis size, the performance decreases. However, this can be avoided with penalizing the trajectory output on convolutional kernels ($[-1,1], [-1,2,-1]$).  \textit{Right:} This  regularization is implicitly achieved by the neural basis. After a certain number of basis functions, the performance remains approximately the same.}
\label{fig:high_pass}
\vspace{-10pt}
\end{figure}

In our architecture also we observed similar behaviors. As shown in the blue curve of the left figure in Fig.~\ref{fig:high_pass}, when the number of basis functions is increased, the loss reaches a minimum, but then increases again. This aligns with the intuition that the motion should be restricted to a low-dimensional manifold of smooth trajectories. Then, we apply 1D convolutions on the DCT trajectories with kernels $[-1,1]$ and $[-1,2,-1]$, and minimize the $\ell_1$ norm on the convolution outputs. These DCT trajectories are then used as the basis functions for modeling the light and density field temporal dynamics. As shown by the orange curve, with this strategy, the performance of the model becomes almost agnostic to the basis size after the minimum. This result is similar to the conclusions of \cite{valmadre2012general}. 

Interestingly, we observed that with the neural basis, this regularization is implicitly achieved; see the right plot of Fig. \ref{fig:high_pass}.  The shown results are for the test set of the ball moves scene. As evident, the performance of the model almost saturates after a certain number of basis functions, eliminating the need for carefully tuning the number of basis functions for each sequence. This result is a powerful indication of the strong architectural bias that stems from the neural networks.

\section{Datasets and evaluation}
\label{suppsec:datasets}

We collect four synthetic scenes and four real-world scenes as our dataset. All the scenes consist of RGB images captured from a single moving camera along with camera poses.   The synthetic scenes are color change, falling and scale, light move, and ball move. The color change scene includes texture changes of objects. The light move scene contains static objects, but a moving light source. The falling and scale scene contains objects that change scale, and the ball move scene consist of objects with long-range movements. Similarly, the real world scenes are climbing, cat walking, flashlight, and flower. The climbing and cat walking scenes contain long-range movements, while the flashlight scene contains light changes. In contrast, the flower scene contains spatially concentrated dynamics.

For each the real world scene, we used $12$ consecutive frames as  training frames, and the subsequent $4$ frames as  testing frames, throughout the video. For the synthetic scenes, we used an unseen fixed pose to render the test frames across time. For evaluation, we used PSNR, SSIM, and LPIPS, as commonly done in literature \cite{tretschk2021non, park2021nerfies, pumarola2021d}.

\section{Hyperparameters and training}
\label{suppsec:hyper}

We use $24$ basis functions for modeling each of the light and density fields. For manifold regularization, we use $8$ as the submanifold dimension. For generating the neural trajectories, we use three-layer ReLU networks  with positional embeddings. We choose $0.1$ for $\lambda_1$ and $\lambda_2$ in Eq.~\ref{eq:loss}. For training, we used an ADAM optimizer with $\beta_1 = 0.9$ and $\beta_2 = 0.99$. We used cyclic learning rates for training both neural networks and the coefficient tensors. For the neural networks, we start the learning rate at $0.001$, and for the coefficient tensors, we start the learning rate at $0.02$. 

\subsection{Volume Rendering}

Let us denote $\mathcal{C}_{\mathbf{x}}(t)$ and $\mathcal{Z}_{\mathbf{x}}(t)$ as continuously evolving light and density fields, respectively,  obtained via Eq.~10 (main paper) and queried at 3D position $\mathbf{x}$.
We can obtain density and light values at any $\mathbf{x}$ at time $t$ as,
  \vspace{-0.5em}
\begin{equation}
    \sigma(\mathbf{x},t), c(\mathbf{x},t) = \mathcal{Z}_{\mathbf{x}}(t), \mathcal{C}_{\mathbf{x}}(t).
\end{equation}

 To compute the above values at an arbitrary continuous position $\mathbf{x}$, we tri-linearly interpolate the grids. Then, the rendering is done similarly to the original NeRF formulation: let $\mathbf{x}(h) = \mathbf{o} + h\mathbf{d}$ be a 3D location sampled on the ray emitted from camera center $\mathbf{o}$ in the direction of $\mathbf{d}$, passing through a pixel $p$. We can obtain the predicted pixel color $\Tilde{p}$ at a given time instance $t$ as,
 \vspace{-0.5em}
\begin{equation}
    \Tilde{p}(t) = \int T(\sigma(\mathbf{x}(h),t), h) \sigma(\mathbf{x}(h),t) c(\mathbf{x}(h),t)dh
\end{equation}
where $T(\cdot) = \exp \big(- \int_{-\infty}^{\mathbf{x}(h)} \sigma(\mathbf{x}(h),t)dh \big)$.
We use the same discrete approximations  used in \cite{mildenhall2021nerf} for the above formulas in practice. The final loss $\mathcal{L}$ used for training is the mean squared loss between $p$ and $\Tilde{p}$, along with a total variation (TV) loss spatially applied across grid values:
 \vspace{-0.5em}
\begin{equation}
    \mathcal{L} = \frac{1}{N}\sum_{i=1}^{N}\|p(t) -  \Tilde{p}(t)\| + \lambda_1 TV(\mathcal{Z}(t)) +  \lambda_2 TV(\mathcal{C}(t)).
\label{eq:loss}
\end{equation}

\section{Convergence}
\begin{figure}[!htp]
\centering
\includegraphics[width=0.6\columnwidth]{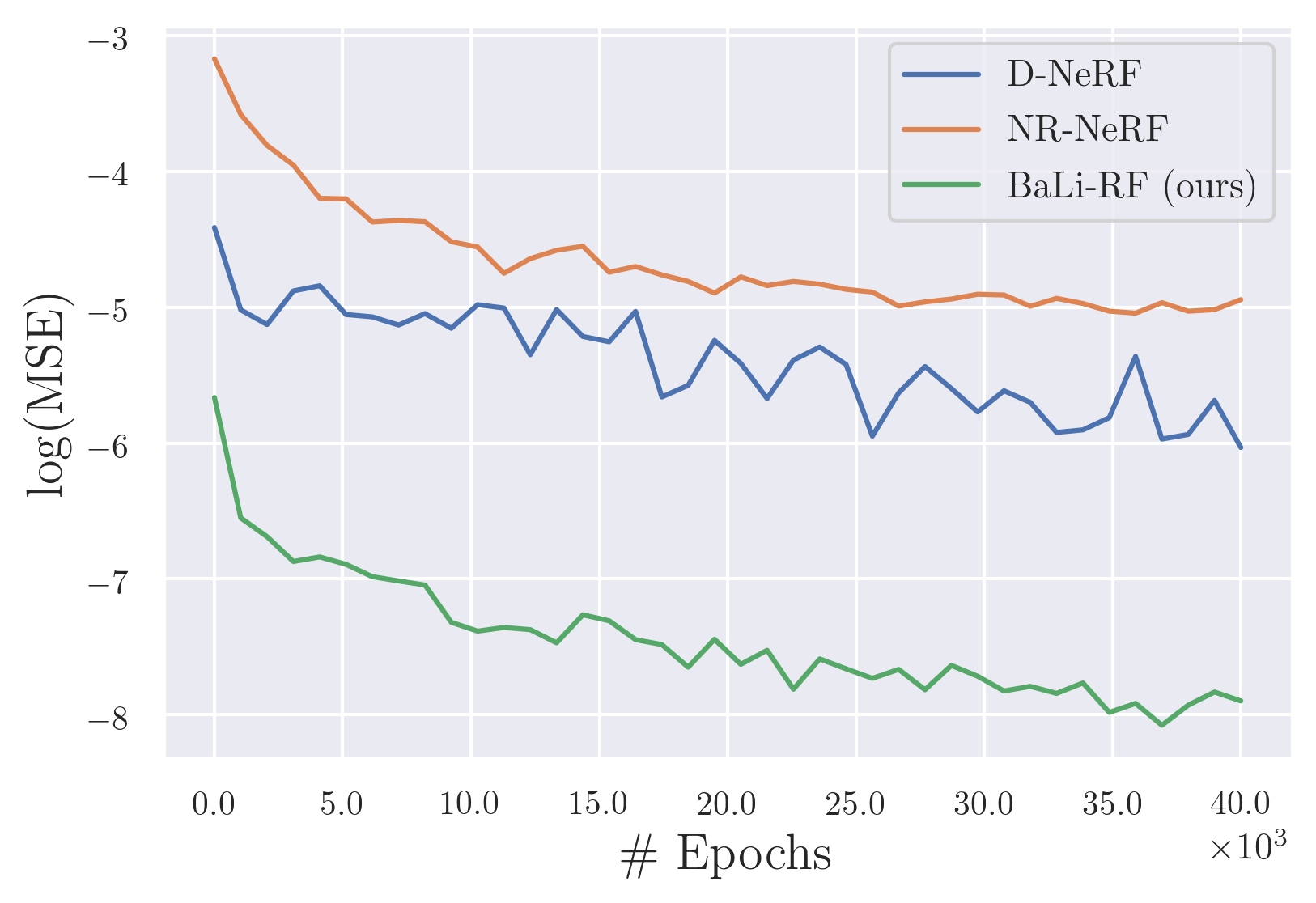}
\vspace{-5pt}
\caption{\textbf{Convergence.} Our model exhibits faster training compared to D-NeRF and NR-NeRF, and converges in $\sim 40k$ epochs. In comparison, D-NeRF and NR-NeRF take $\sim 800k$ and $\sim 200k$ epochs to converge, respectively. Time-wise, our model trains in $\sim 1.5$ hours per a scene, which is  $\sim 20\times$ faster and $\sim10 \times$ faster compared to D-NeRF and NR-NeRF, respectively.}
\vspace{-5pt}
\label{fig:convergence}
\end{figure}

\vspace{-1em}
Our method converges around $20 \times$ faster than D-NeRF and $10 \times$ faster compared to NR-NeRF (Fig.~\ref{fig:convergence}). Also, we noticed that our convergence is more stable compared the baselines. For instance, NR-NeRF exhibited sudden divergences from the minima when the training is continued for a long time. Therefore, it was necessary to carefully monitor the training to determine the optimal termination point.


\vspace{-0.5em}

\section{T-TensoRF}
\label{suppsec:tnerf}

\begin{figure}[!htp]
\centering
\includegraphics[width=1.0\columnwidth]{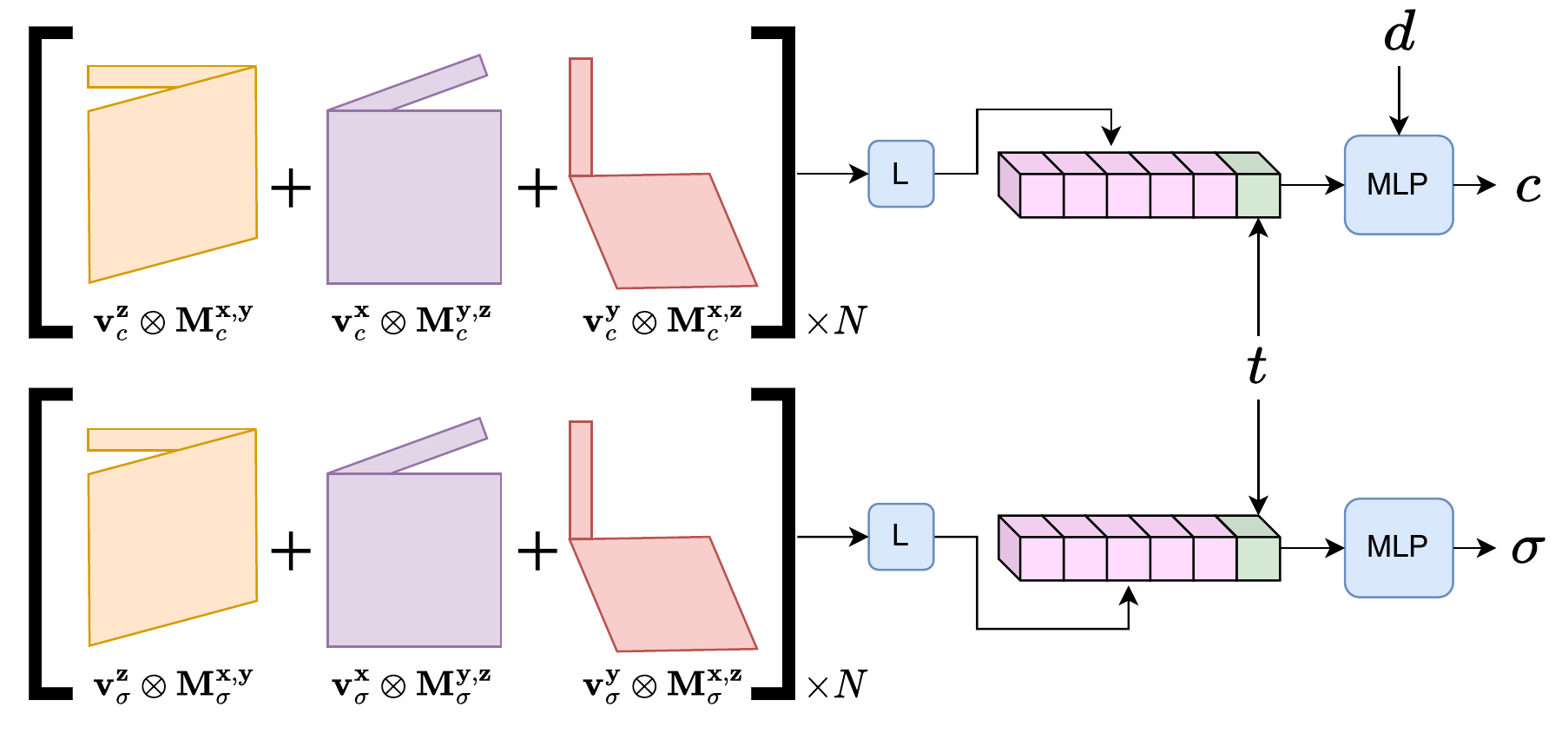}
\caption{\textbf{The T-TensoRF architecture.} We develop a baseline that disentangles light and density fields, but does not factorize time and space. See Sec.~\ref{suppsec:tnerf} for a detailed description.}
\label{fig:t-tensorf}
\end{figure}

Two unique features of our framework are the light/density disentanglement and the space/time factorization. Thus, it is necessary to properly evaluate the superior performance of our model against these two factors. To this end, we design a baseline which completely disentangles the light and density fields, but does not factorize space and time. Fig. \ref{fig:t-tensorf} shows the overall architecture. Here, we first model the light $\mathcal{S}_{c}$  and density $\mathcal{S}_{\sigma}$  fields as 3D tensors, which is decomposed in to a linear combination of outer products between matrices and vectors:

\begin{equation}
  \mathcal{S}_{\sigma} = \sum_{j=1}^N( \mathbf{v}^z_{\sigma, j} \otimes \mathbf{M}_{\sigma, j}^{xy} + \mathbf{v}^x_{\sigma, j} \otimes \mathbf{M}_{\sigma, j}^{yz} + \mathbf{v}^y_{\sigma, j} \otimes \mathbf{M}_{\sigma, j}^{xz}),
\label{eq:t-nerf-density}
\end{equation}

\begin{equation}
  \mathcal{S}_{c} = \sum_{j=1}^N( \mathbf{v}^z_{c, j} \otimes \mathbf{M}_{c, j}^{xy} + \mathbf{v}^x_{c, j} \otimes \mathbf{M}_{c, j}^{yz} + \mathbf{v}^y_{c, j} \otimes \mathbf{M}_{c, j}^{xz}),
\label{eq:t-nerf-color}
\end{equation}

For querying continuous 3D positions, we tri-linearly interpolate the resultant grid. Let

\begin{equation}
    R_{c,j} = ( \mathbf{v}^z_{c, j} \otimes \mathbf{M}_{c, j}^{xy} + \mathbf{v}^x_{c, j} \otimes \mathbf{M}_{c, j}^{yz} + \mathbf{v}^y_{c, j} \otimes \mathbf{M}_{c, j}^{xz})
\end{equation}

and

\begin{equation}
   R_{\sigma,j} =  ( \mathbf{v}^z_{\sigma, j} \otimes \mathbf{M}_{\sigma, j}^{xy} + \mathbf{v}^x_{\sigma, j} \otimes \mathbf{M}_{\sigma, j}^{yz} + \mathbf{v}^y_{\sigma, j} \otimes \mathbf{M}_{\sigma, j}^{xz}),
\end{equation}

and $R_{c,j}(\mathbf{x})$, $R_{\sigma,j}(\mathbf{x})$ denote the values queried at $\mathbf{x}$. Then, we use two linear networks $L_\sigma, L_c:\mathbb{R}^N \to \mathbb{R}^F$ to generate $F$-dimensional density/light feature vectors $(\mu_\sigma, \mu_c)$ for each 3D position $\mathbf{x}$ as

\begin{equation}
    \mu_{\sigma} = L_{\sigma}(R_{\sigma,1}, \dots R_{\sigma,N}),
\end{equation}

\begin{equation}
    \mu_c = L_c(R_{c,1}, \dots R_{c,N}).
\end{equation}

This is equivalent to generating a density/light feature vector for each 3D position of the scene. Then, we concatenate these feature vectors with the scalar time value, and feed to a 4-layer ReLU network to obtain color and density values for each $\mathbf{x}$. The neural rendering and training is done similar to our model.

\section{Experiments on the Nvidia dynamic dataset}

we conduct experiments on the NVIDIA dynamic view synthesis dataset \cite{gao2021dynamic}. We report the PSNR and LPIPS in Table \ref{tab:nvidia}. Our method achieves  state-of-the-art results.

\begin{table*}[!htbp]
    \centering
    \footnotesize
    \ra{1.2}
    \renewcommand\tabcolsep{5.5pt}
    \begin{tabularx}{1\textwidth}{l ccccccc|c}
        \toprule
        PSNR $\uparrow$ / LPIPS $\downarrow$ & Jumping & Skating & Truck & Umbrella & Balloon1 & Balloon2 & Playground & Average \\
        \midrule
        NeRF~\cite{mildenhall2021nerf} & 
        20.99 / 0.305 &
        23.67 / 0.311 &
        22.73 / 0.229 &
        21.29 / 0.440 &
        19.82 / 0.205 &
        24.37 / 0.098 &
        21.07 / 0.165 &
        21.99 / 0.250 \\
        D-NeRF~\cite{pumarola2021d} & 
        22.36 / 0.193 & 
        22.48 / 0.323 & 
        24.10 / 0.145 & 
        21.47 / 0.264 & 
        19.06 / 0.259 & 
        20.76 / 0.277 & 
        20.18 / 0.164 &
        21.48 / 0.232 \\
        NR-NeRF~\cite{tretschk2021non} & 
        20.09 / 0.287 & 
        23.95 / 0.227 & 
        19.33 / 0.446 & 
        19.63 / 0.421 & 
        17.39 / 0.348 & 
        22.41 / 0.213 & 
        15.06 / 0.317 &
        19.69 / 0.323 \\
        NSFF~\cite{li2021neural} & 
        24.65 / 0.151 & 
        29.29 / 0.129 & 
        25.96 / 0.167 & 
        22.97 / 0.295 & 
        21.96 / 0.215 & 
        24.27 / 0.222 & 
        21.22 / 0.212 &
        24.33 / 0.199 \\
        DynamicNeRF~\cite{gao2021dynamic} & 
        24.68 / 0.090 & 
        \textbf{32.66} / \textbf{0.035} & 
        28.56 / 0.082 & 
        23.26 / 0.137 &
        22.36 / 0.104 & 
        27.06 / \textbf{0.049} & 
        24.15 / 0.080 &
        26.10 / 0.082 \\
        HyperNeRF~\cite{park2021hypernerf} & 
        18.34 / 0.302 &
        21.97 / 0.183 & 
        20.61 / 0.205 & 
        18.59 / 0.443 & 
        13.96 / 0.530 & 
        16.57 / 0.411 & 
        13.17 / 0.495 &
        17.60 / 0.367 \\
        TiNeuVox~\cite{TiNeuVox} & 
        20.81 / 0.247 &
        23.32 / 0.152 & 
        23.86 / 0.173 & 
        20.00 / 0.355 & 
        17.30 / 0.353 & 
        19.06 / 0.279 & 
        13.84 / 0.437 &
        19.74 / 0.285 \\
        RoDynRF~\cite{liu2023robust} &  
        25.66 / \textbf{0.071} & 
        28.68 / 0.040 & 
        29.13 / \textbf{0.063} & 
        24.26 / \textbf{0.089} & 
        \textbf{22.37} / \textbf{0.103} & 
        26.19 / 0.054 & 
        \textbf{24.96} / \textbf{0.048} & 
        25.89 / \textbf{0.065} \\
        \ours (ours) &
        \textbf{27.55} / 0.145 &
        32.00 / 0.106 &
        \textbf{29.99} / 0.206 &
        \textbf{26.32} / 0.207 &
        22.22 / 0.263 &
        \textbf{28.02} / 0.173 &
        23.96 / 0.136 &
        \textbf{27.15} / 0.177  \\
        \bottomrule
    \end{tabularx}
     \vspace{-0.5em}
    \caption{\textbf{Comparison on the Nvidia Dynamic Scene dataset.} Numbers for the competing methods are extracted from \cite{liu2023robust}.}
    \label{tab:nvidia}
\end{table*}
 \vspace{-0.5em}

\section{Novel view generation}
\label{suppsec:comparisons}

\begin{figure}[!htp]
\centering
\includegraphics[width=0.7\columnwidth]{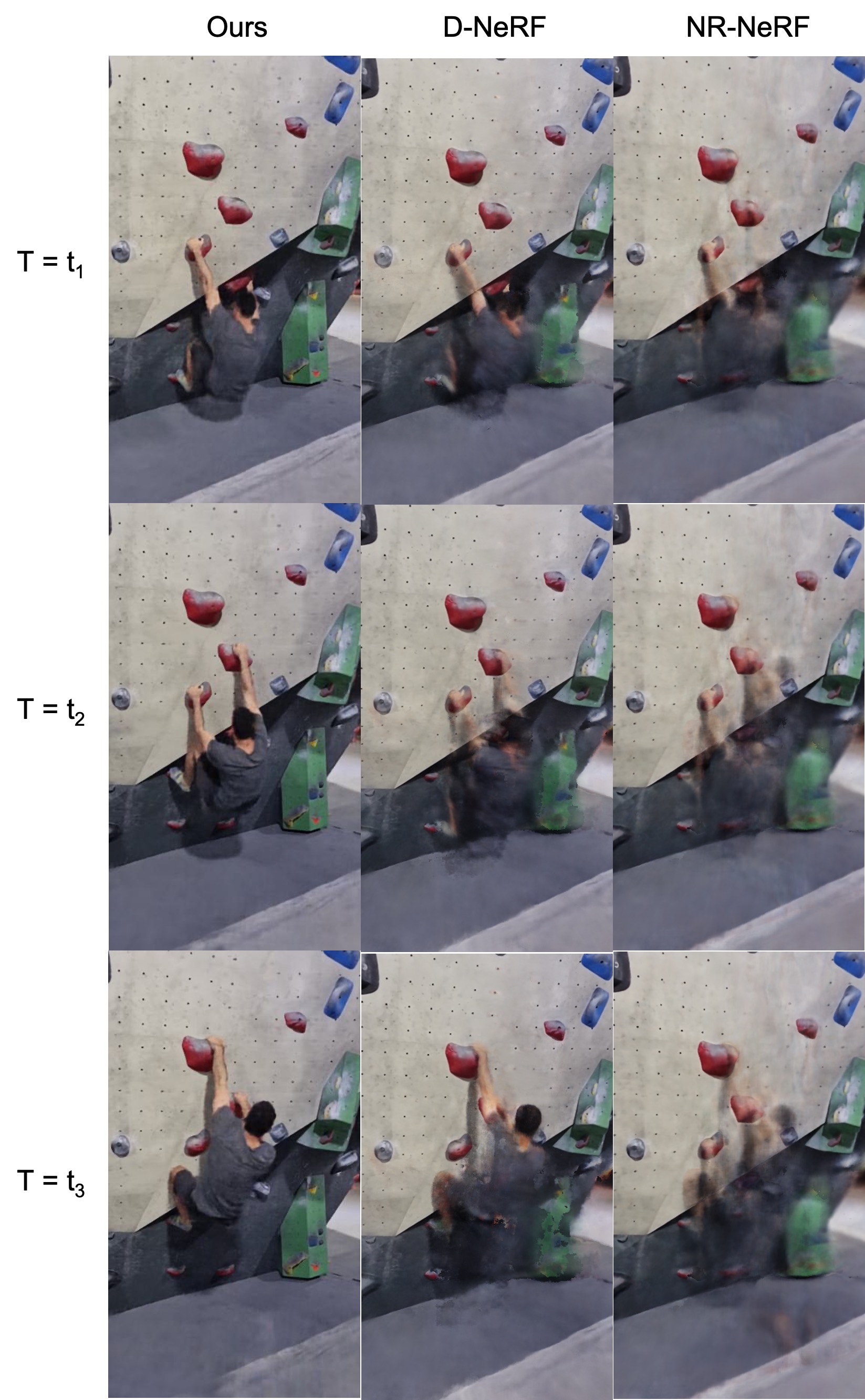}
\caption{\textbf{A qualitative comparison over the generated novel views on the climbing scene.} We fix the pose and generate views by varying time. As depicted, our model is able to achieve superiors results in all the instances.}
\label{fig:climbing_fp}
\vspace{-10pt}
\end{figure}

\begin{figure}[!htp]
\centering
\includegraphics[width=1.\columnwidth]{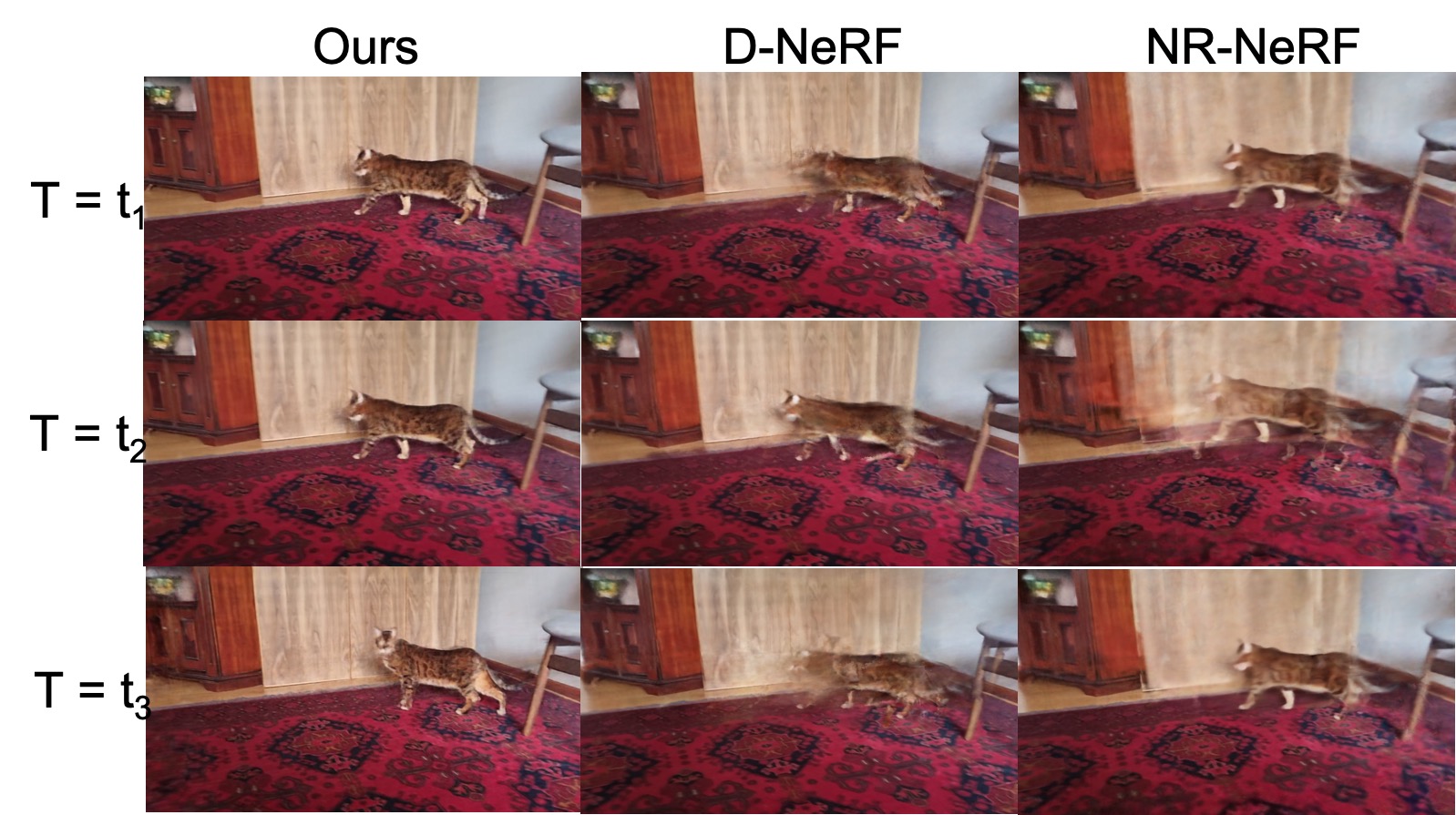}
\caption{\textbf{A qualitative comparison over the generated novel views on the cat scene.} We fix the pose and generate views by varying time. As depicted, our model is able to achieve superiors results in all the instances.}
\label{fig:cat_fp}
\vspace{-10pt}
\end{figure}

\begin{figure}[!htp]
\centering
\includegraphics[width=1.\columnwidth]{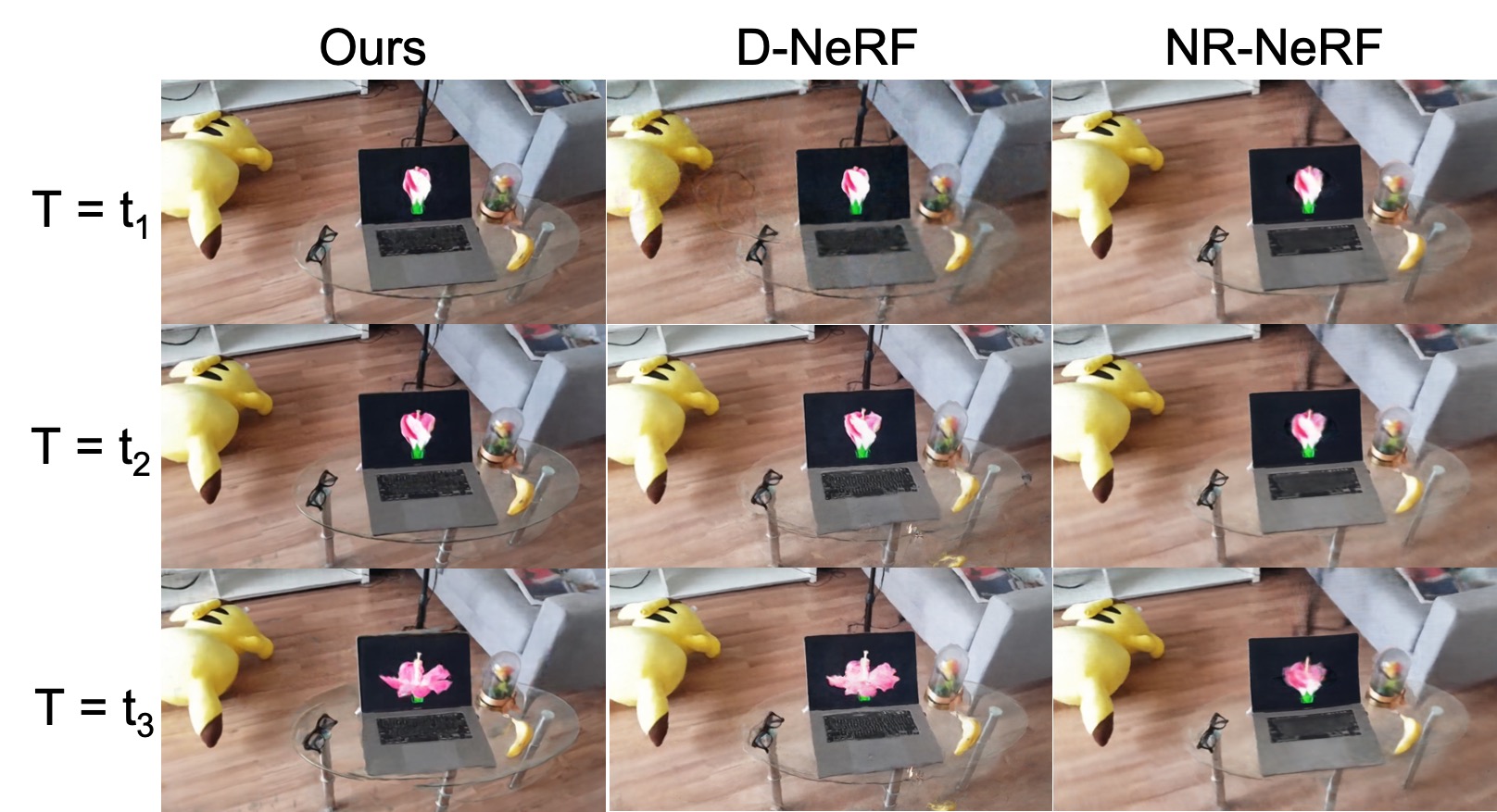}
\caption{\textbf{A qualitative comparison over the generated novel views on the flower scene.} We fix the pose and generate views by varying time. As depicted, our model is able to achieve superiors results in all the instances.}
\label{fig:flower_fp}
\vspace{-10pt}
\end{figure}

\begin{figure}[!htp]
\centering
\includegraphics[width=1.\columnwidth]{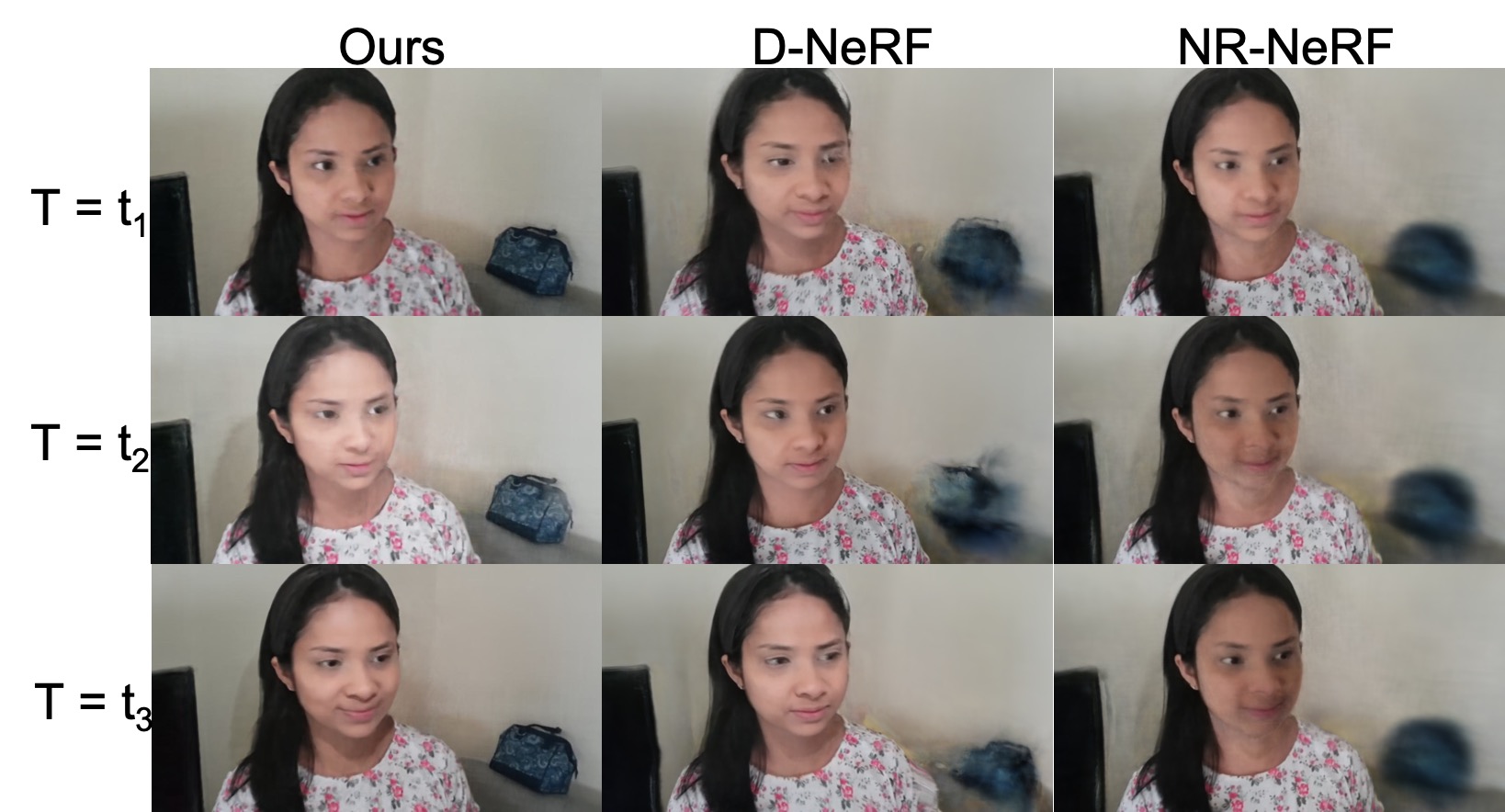}
\caption{\textbf{A qualitative comparison over the generated novel views on the flashlight scene.} We fix the pose and generate views by varying time. As depicted, our model is able to achieve superiors results in all the instances.}
\label{fig:flashlight_fp}
\vspace{-10pt}
\end{figure}

\begin{figure}[!htp]
\centering
\includegraphics[width=1.\columnwidth]{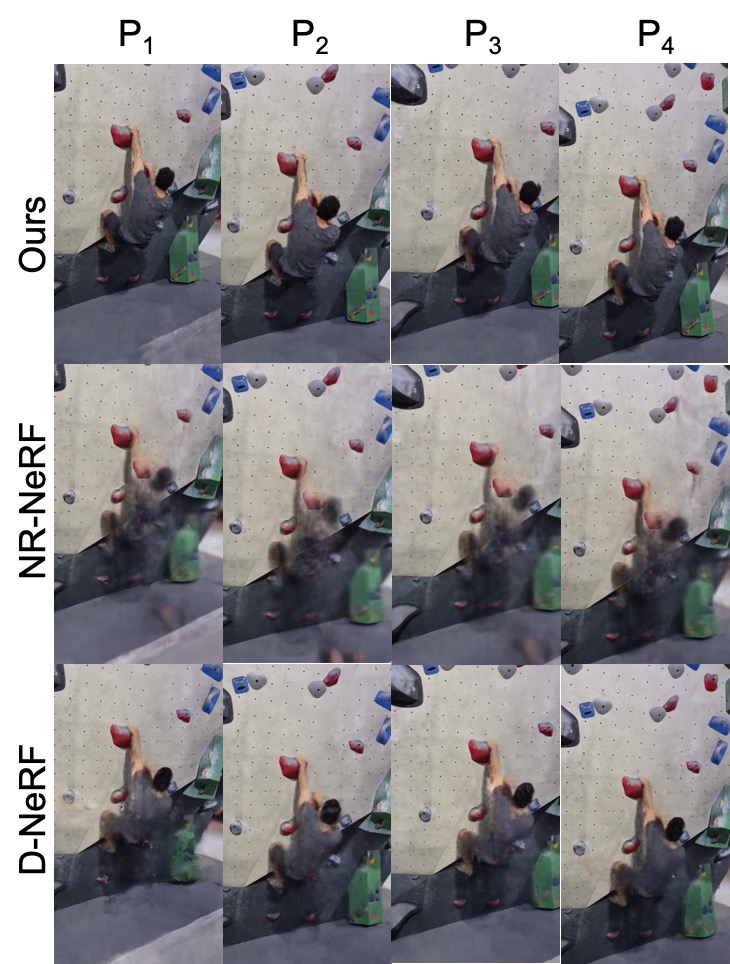}
\caption{\textbf{A qualitative comparison over the generated novel views on the climbing scene.} We fix the time and generate views from different camera poses. As depicted, our model is able to achieve superiors results in all the instances.}
\label{fig:climbing_ft}
\vspace{-10pt}
\end{figure}

\begin{figure}[!htp]
\centering
\includegraphics[width=1.\columnwidth]{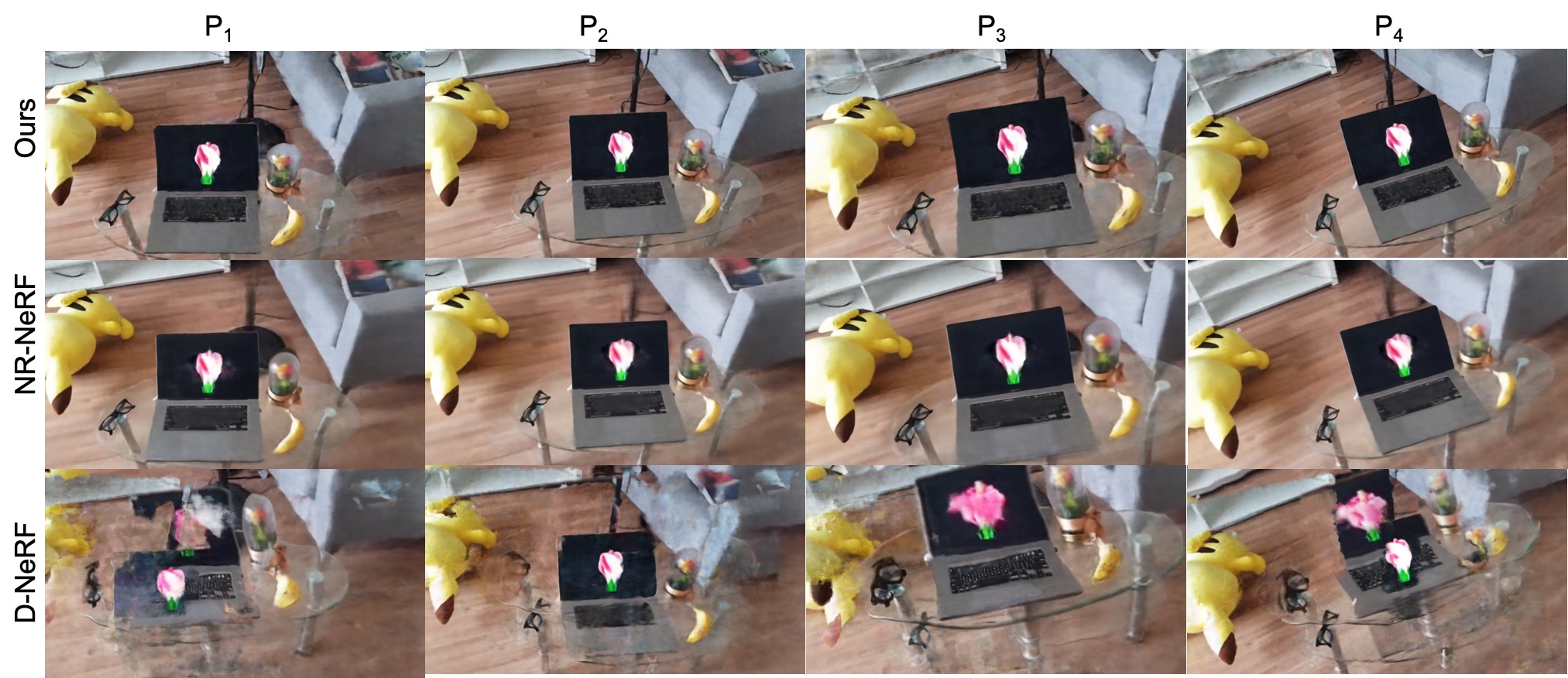}
\caption{\textbf{A qualitative comparison over the generated novel views on the flower scene.} We fix the time and generate views from different camera poses. As depicted, our model is able to achieve superiors results in all the instances.}
\label{fig:flower_ft}
\end{figure}

\begin{figure}[!htp]
\centering
\includegraphics[width=1.\columnwidth]{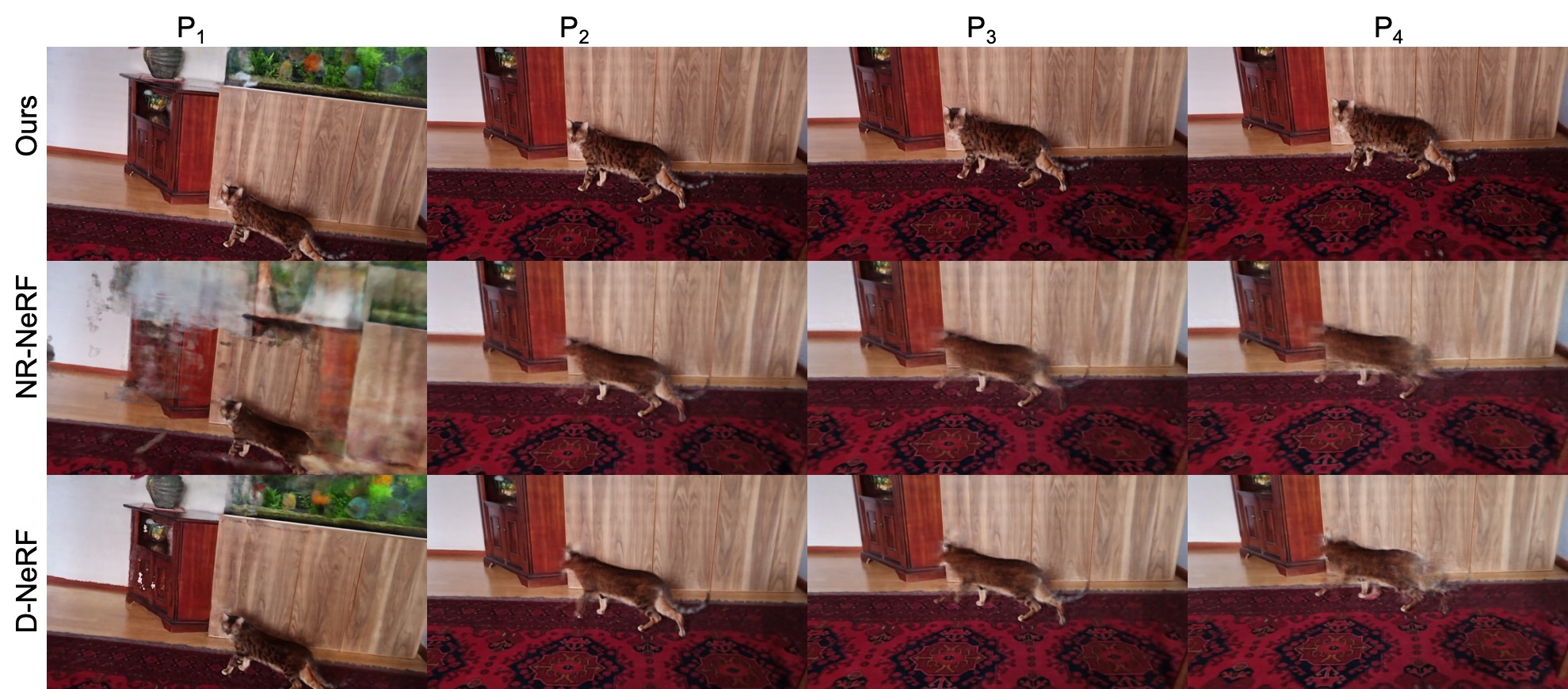}
\caption{\textbf{A qualitative comparison over the generated novel views on the cat scene.} We fix the time and generate views from different camera poses. As depicted, our model is able to achieve superiors results in all the instances.}
\label{fig:cat_ft}
\end{figure}

\begin{figure}[!htp]
\centering
\includegraphics[width=1.\columnwidth]{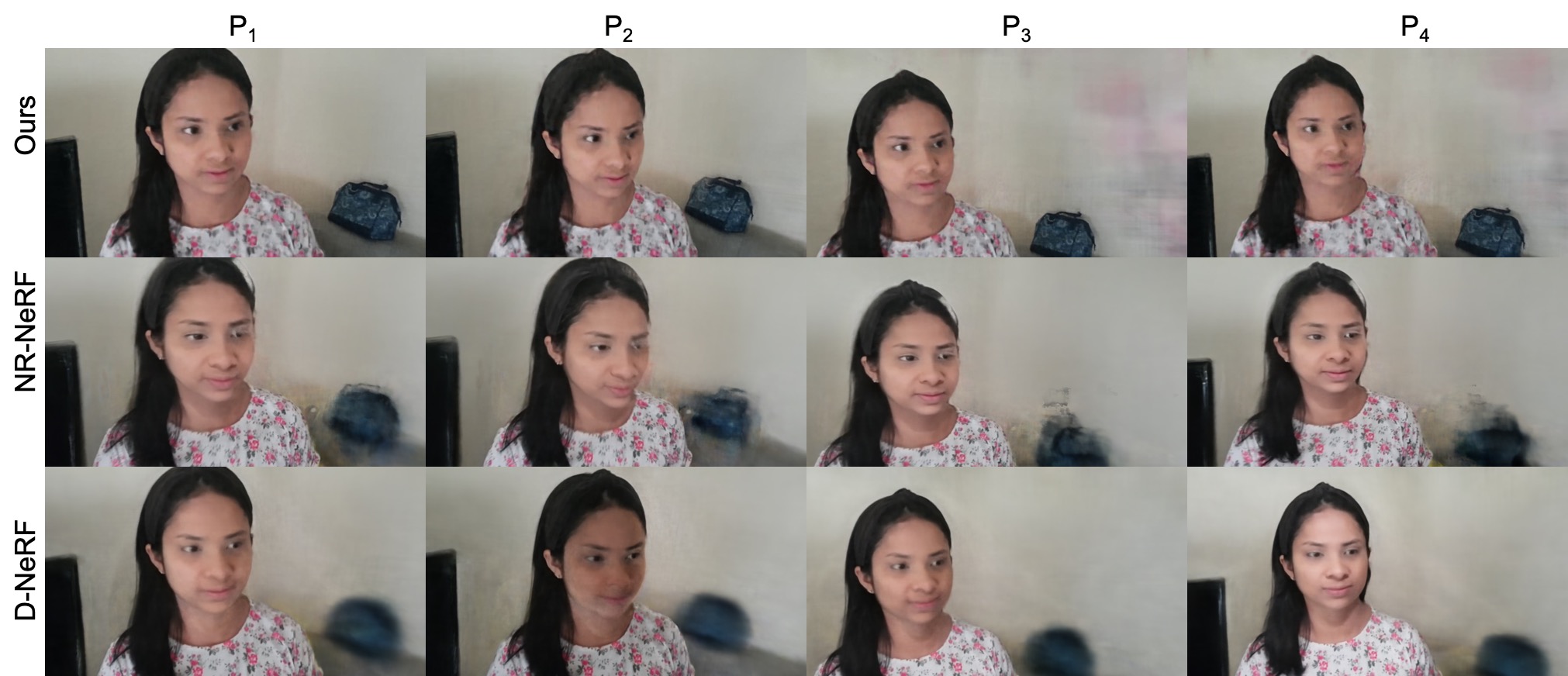}
\caption{\textbf{A qualitative comparison over the generated novel views on the flashlight scene.} We fix the time and generate views from different camera poses. As depicted, our model is able to achieve superiors results in all the instances.}
\label{fig:flashlight_ft}
\end{figure}

In this section, we offer more qualitative comparisons. For real world scenes, we first fix the pose and move time to generate novel views. Fig.~\ref{fig:climbing_fp}, \ref{fig:cat_fp}, \ref{fig:flower_fp}, and \ref{fig:flashlight_fp} depict results. Next, we fix the time and generate novel views by changing the poses. Fig.~\ref{fig:climbing_ft}, \ref{fig:cat_ft}, \ref{fig:flower_ft}, and \ref{fig:flashlight_ft} depict corresponding results. As evident, our model exhibits significantly superior performance over all the instances. Recall that the training images for these scenes are obtained from a single moving camera. Therefore, only a single image is available for a particular time instance. Thus, this reconstruction task is a severely underconstrained problem, specially in the context of complex real world dynamics. Therefore, the superior results shown by our model is a strong indicator of its inbuilt architectural bias that implicitly regularizes the problem. 

\begin{figure}[!htp]
\centering
\includegraphics[width=1.\columnwidth]{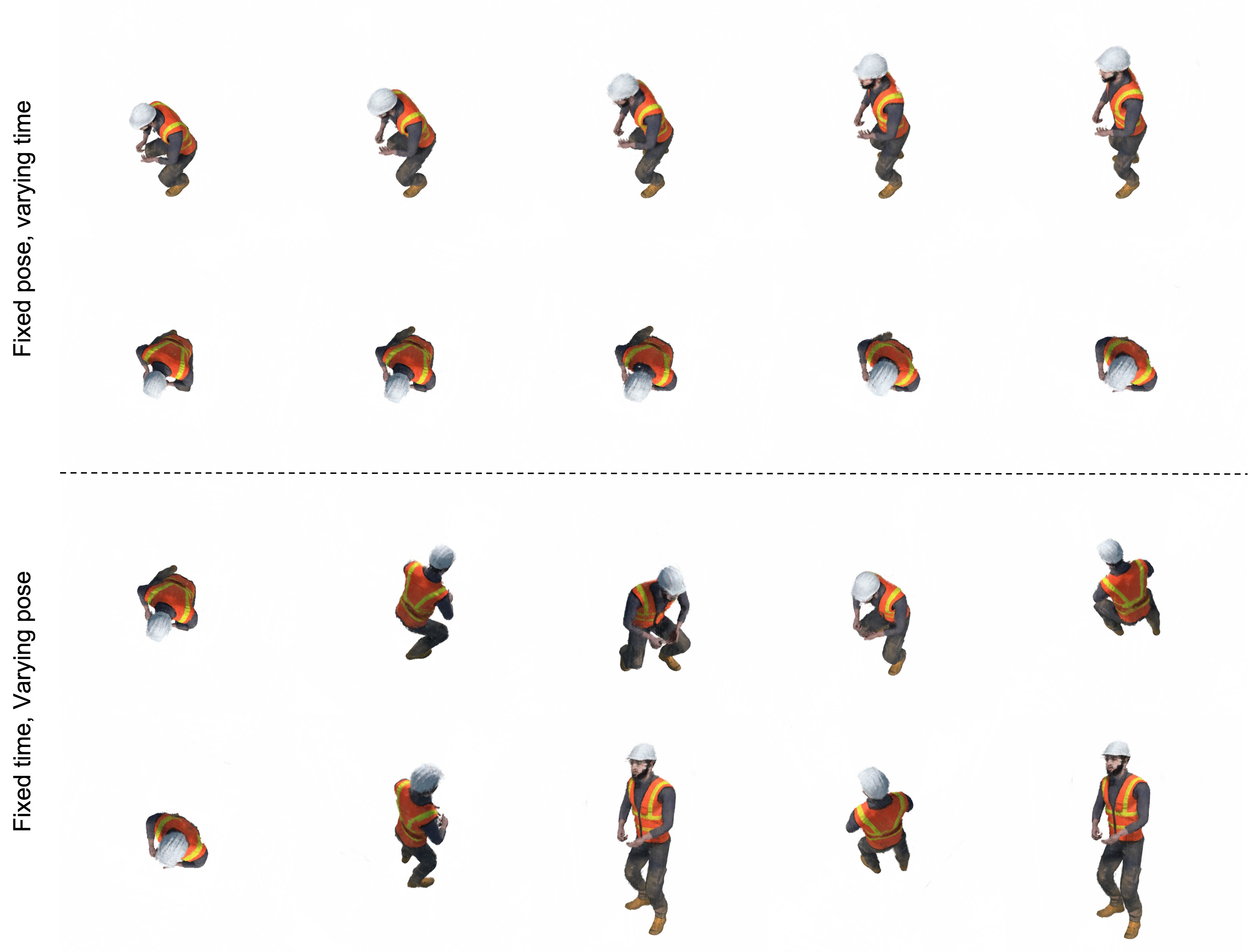}
\caption{Qualitative examples generated by our model on the \emph{standup} scene in \cite{pumarola2021d} synthetic dataset.}
\label{fig:standup}
\vspace{-10pt}
\end{figure}

\begin{figure}[!htp]
\centering
\includegraphics[width=1.\columnwidth]{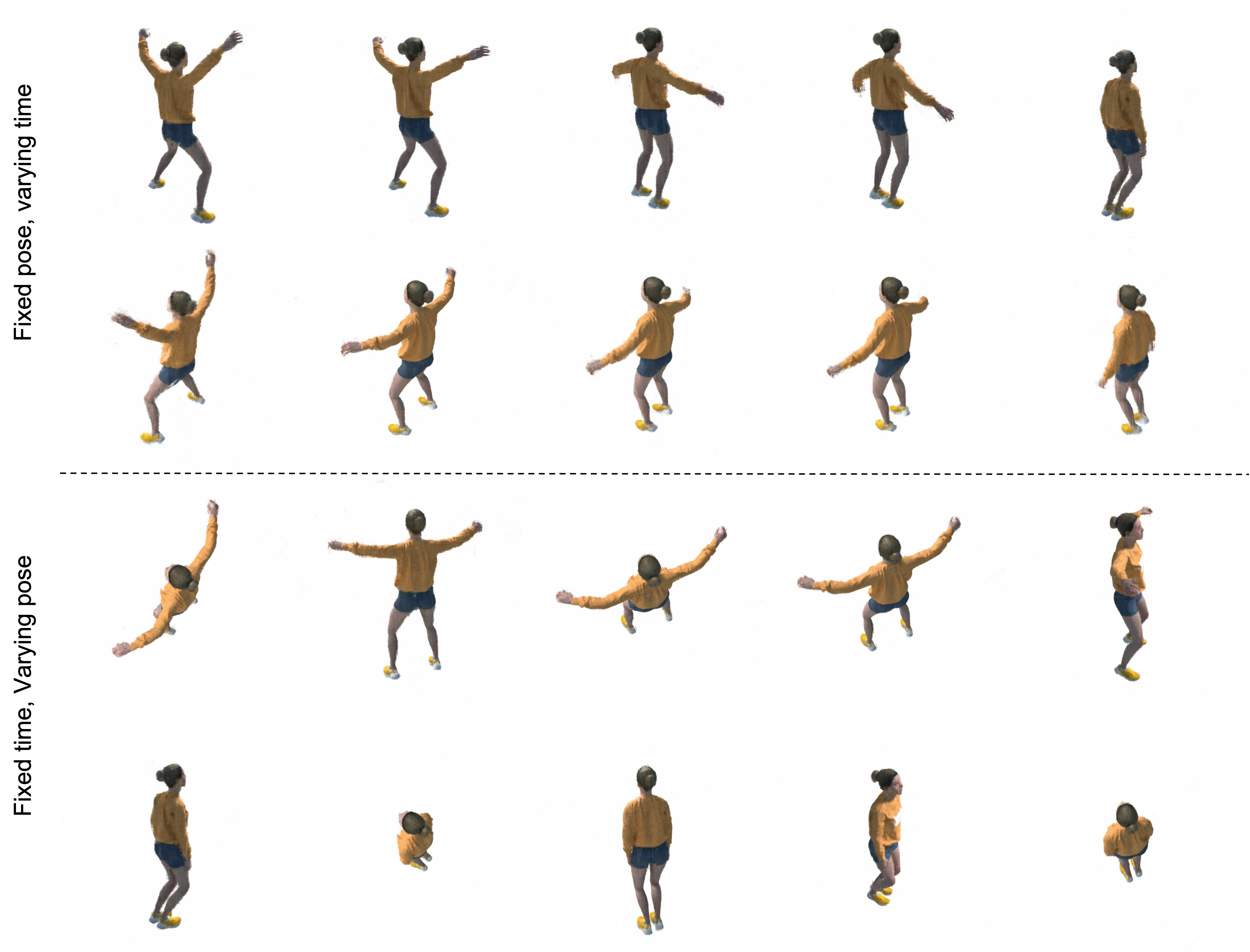}
\caption{Qualitative examples generated by our model on the \emph{jumping} scene in \cite{pumarola2021d} synthetic dataset.}
\label{fig:jumping}
\vspace{-10pt}
\end{figure}

\begin{figure}[!htp]
\centering
\includegraphics[width=1.\columnwidth]{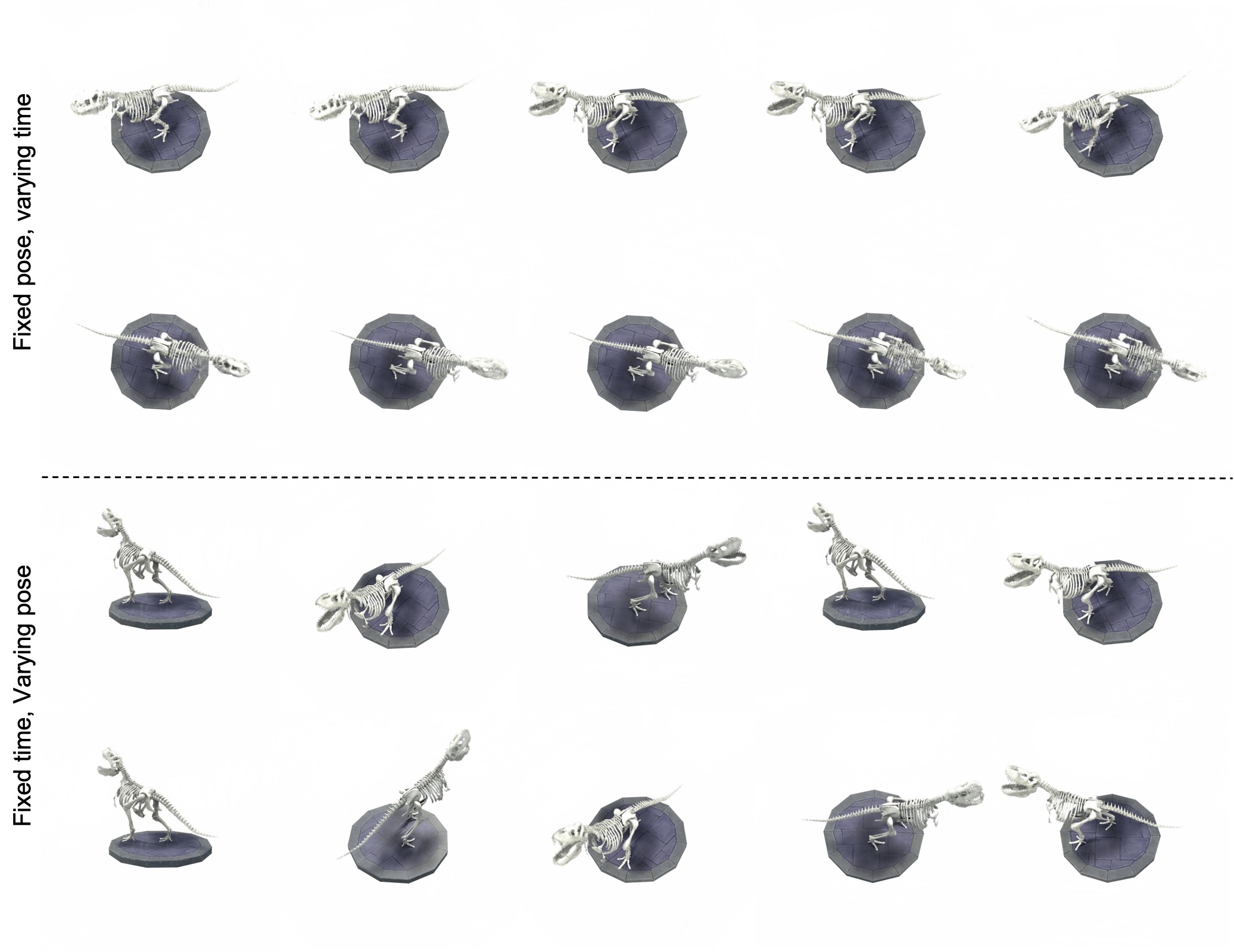}
\caption{Qualitative examples generated by our model on the \emph{trex} scene in \cite{pumarola2021d} synthetic dataset.}
\label{fig:trex}
\vspace{-10pt}
\end{figure}

\begin{figure}[!htp]
\centering
\includegraphics[width=1.\columnwidth]{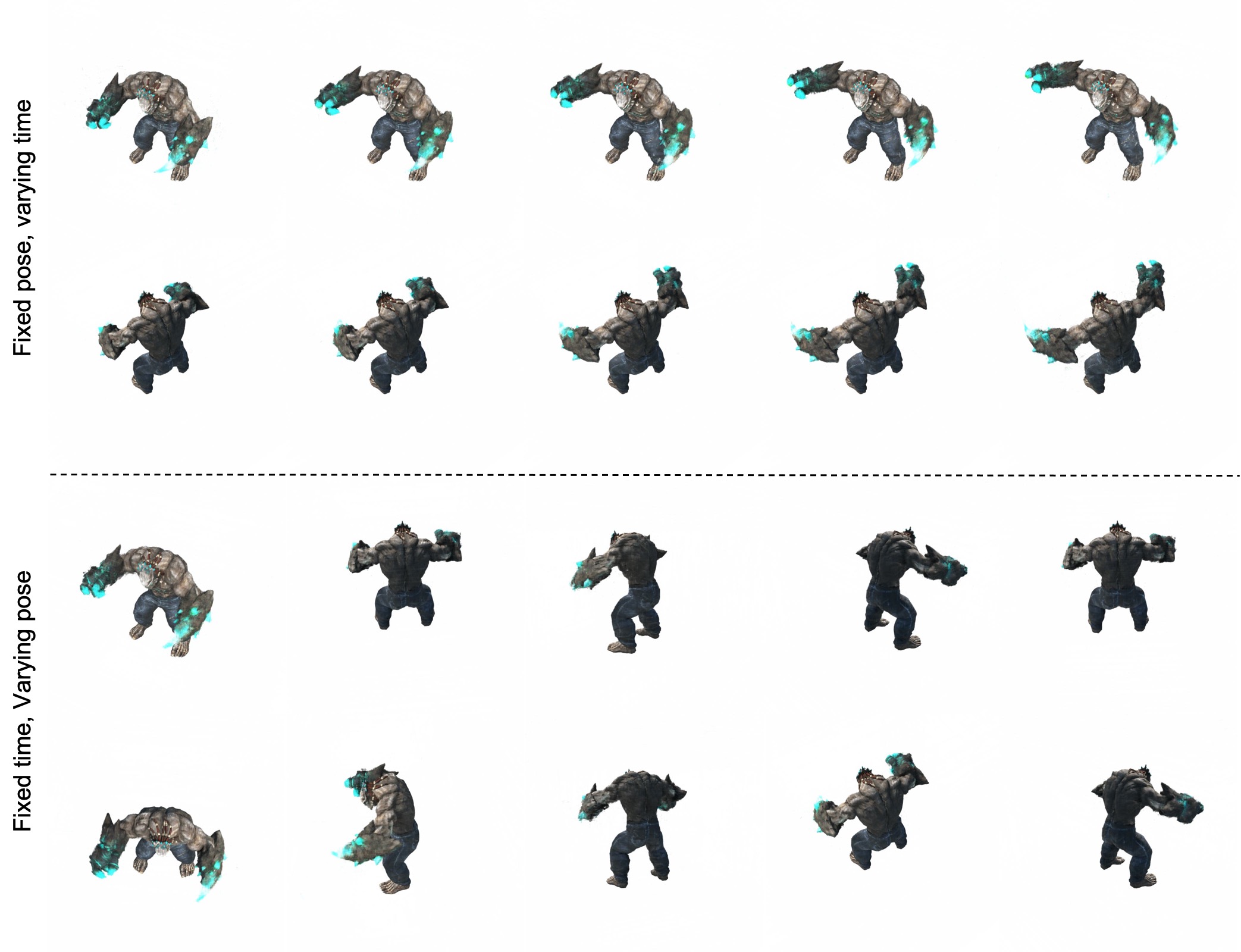}
\caption{Qualitative examples generated by our model on the \emph{mutant} scene in \cite{pumarola2021d} synthetic dataset.}
\label{fig:mutant}
\vspace{-10pt}
\end{figure}

\begin{figure}[!htp]
\centering
\includegraphics[width=1.\columnwidth]{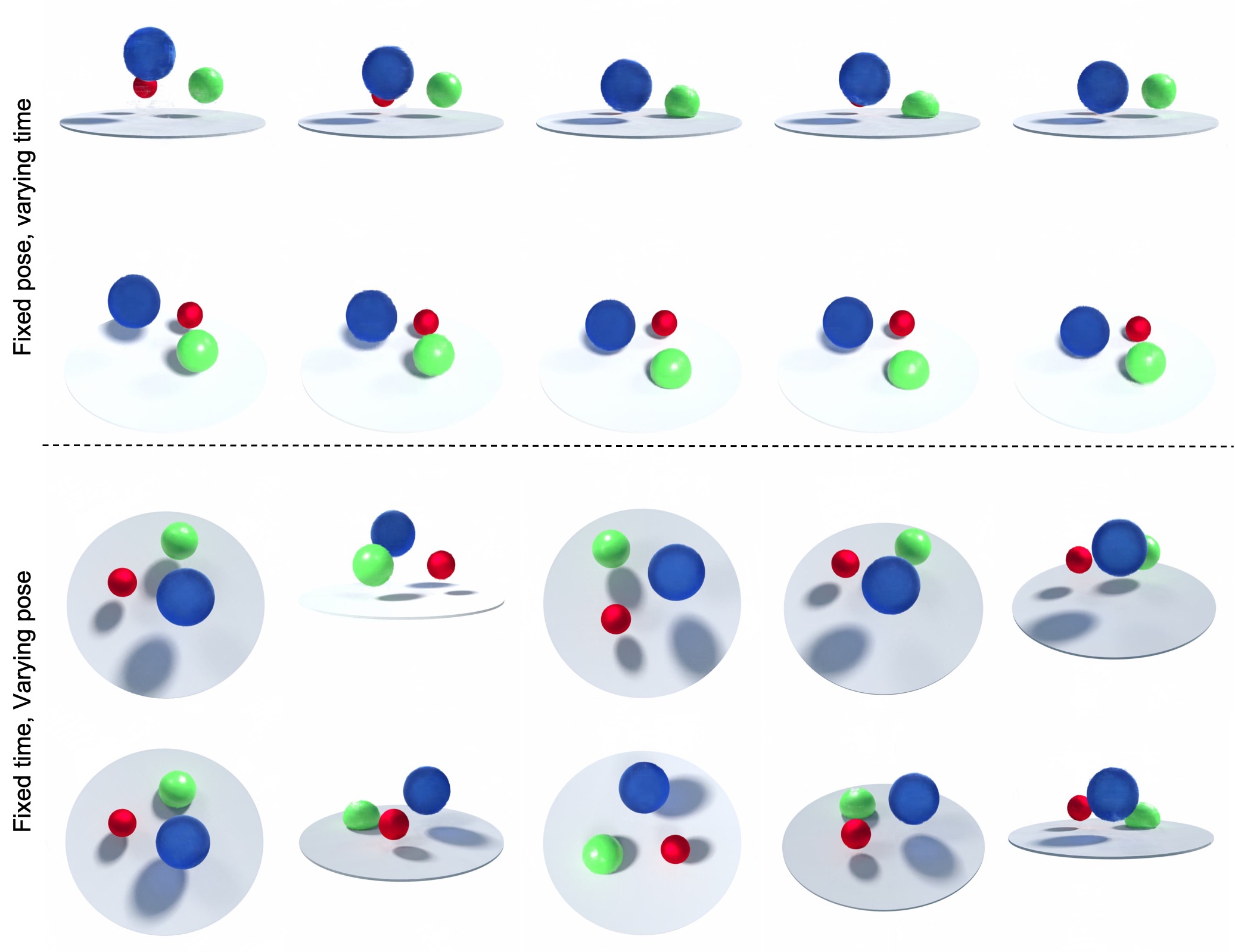}
\caption{Qualitative examples generated by our model on the \emph{bouncing balls} scene in \cite{pumarola2021d} synthetic dataset.}
\label{fig:bb}
\end{figure}

\begin{figure}[!htp]
\centering
\includegraphics[width=1.\columnwidth]{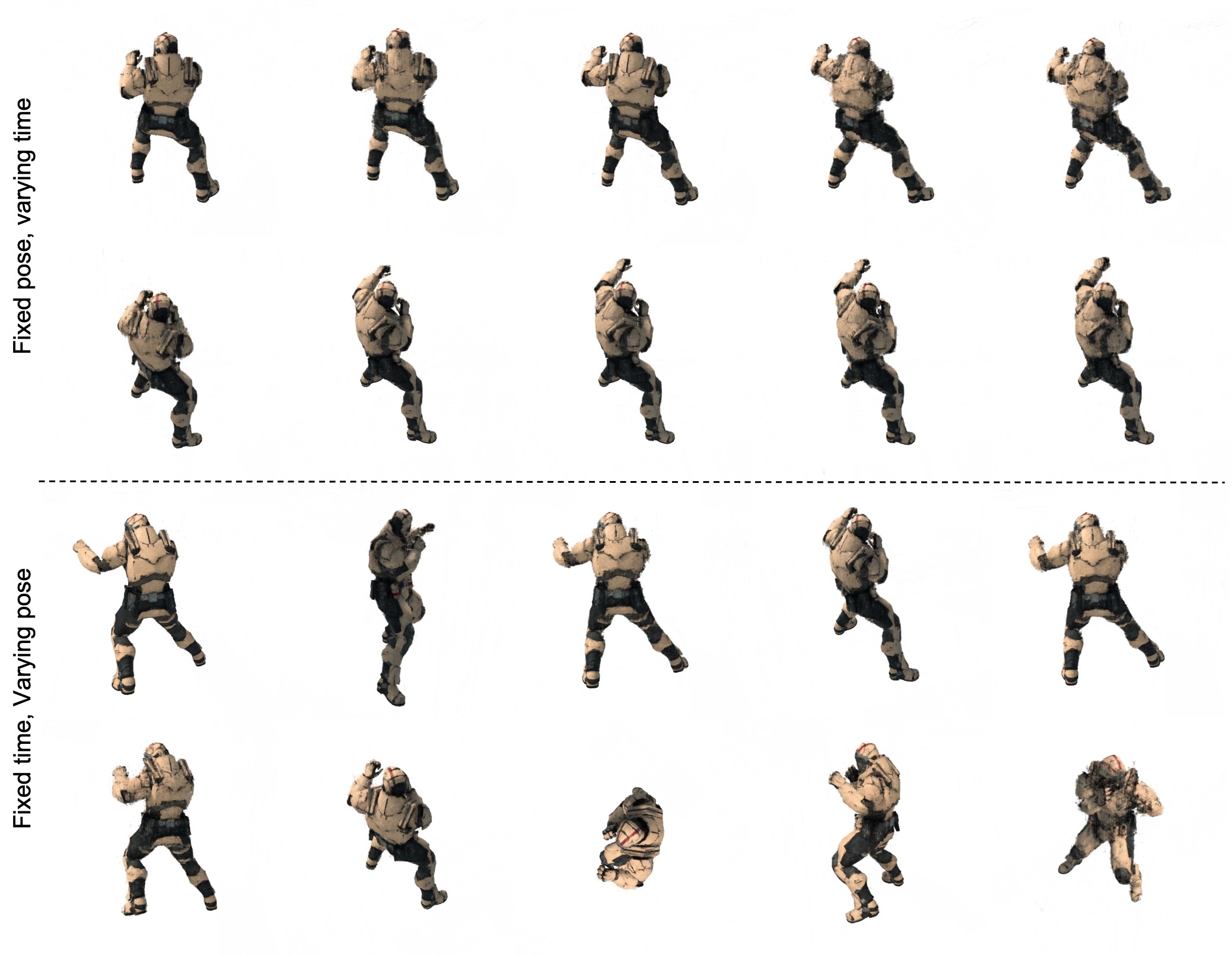}
\caption{Qualitative examples generated by our model on the \emph{hook} scene in \cite{pumarola2021d} synthetic dataset.}
\label{fig:hook_ft}
\end{figure}

We also conduct experiments over the synthetic dataset released by \cite{pumarola2021d}. Fig.~\ref{fig:standup}, \ref{fig:jumping}, \ref{fig:mutant}, \ref{fig:bb}, \ref{fig:trex}, and \ref{fig:hook_ft} depict results. As shown, our model is able to generate novel views in both constant pose and constant time settings. 

\begin{figure}[!htp]
\centering
\includegraphics[width=1.\columnwidth]{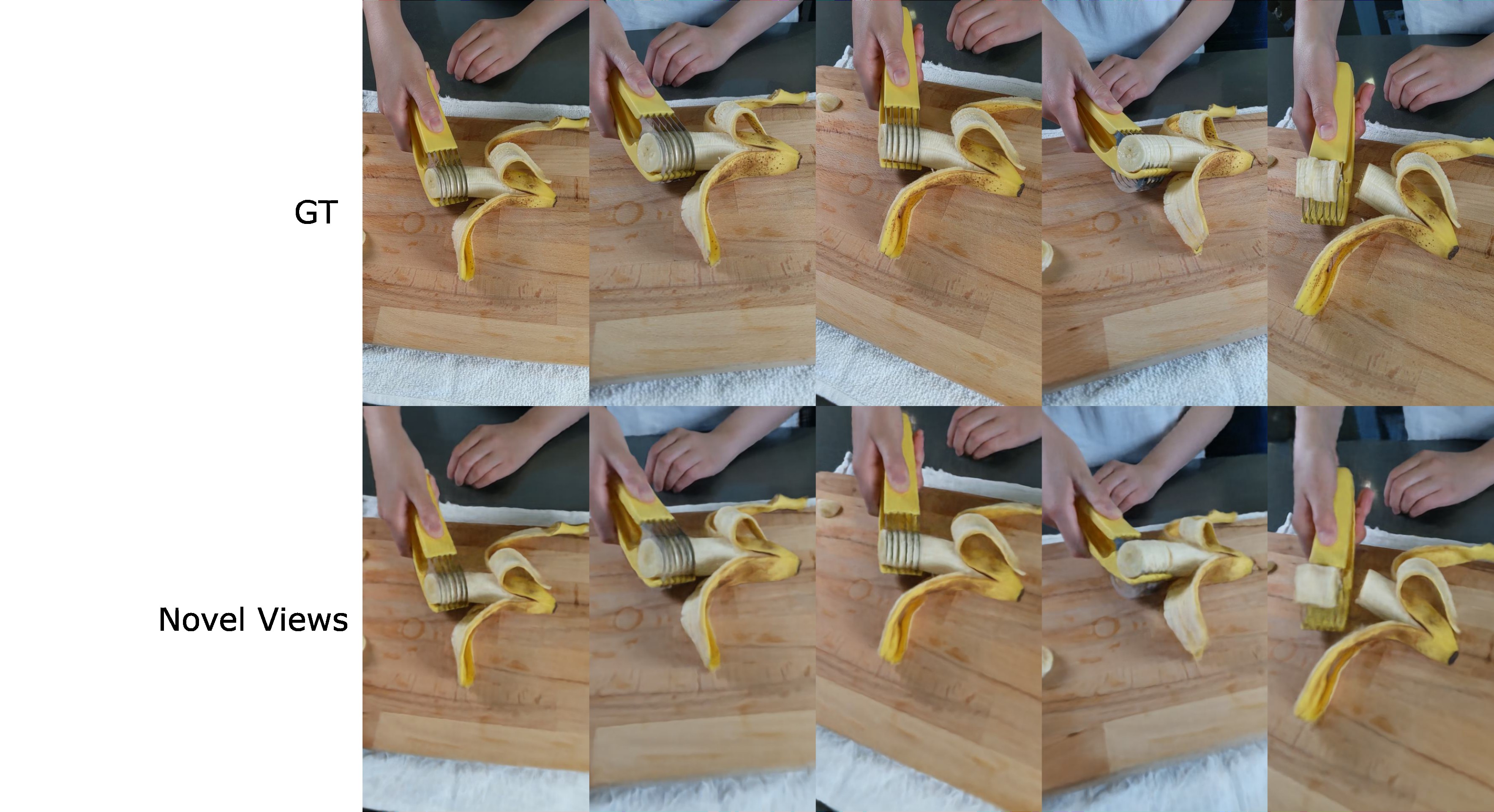}
\caption{Interpolated novel views on the HyperNeRF Banana sequence.}
\label{fig:banana}
\end{figure}

\begin{figure}[!htp]
\centering
\includegraphics[width=1.\columnwidth]{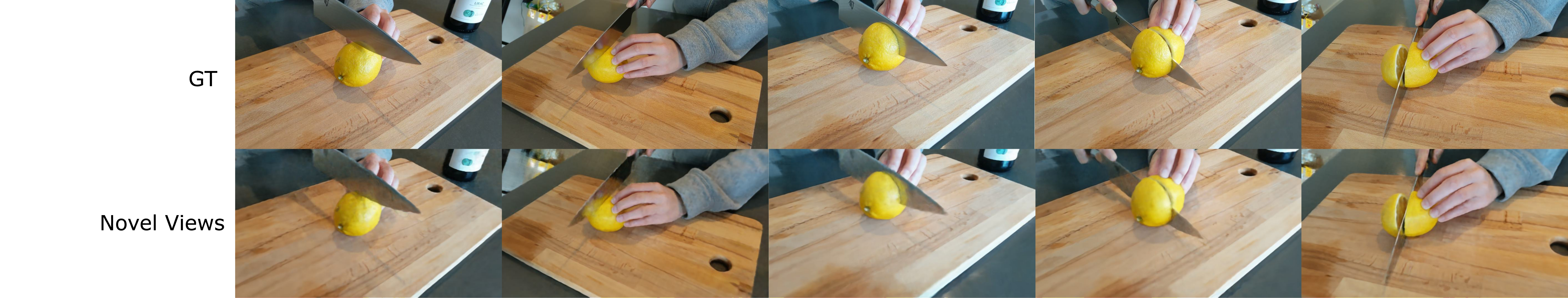}
\caption{Interpolated novel views on the HyperNeRF Lemon sequence.}
\label{fig:lemon}
\end{figure}

\begin{figure}[!htp]
\centering
\includegraphics[width=1.\columnwidth]{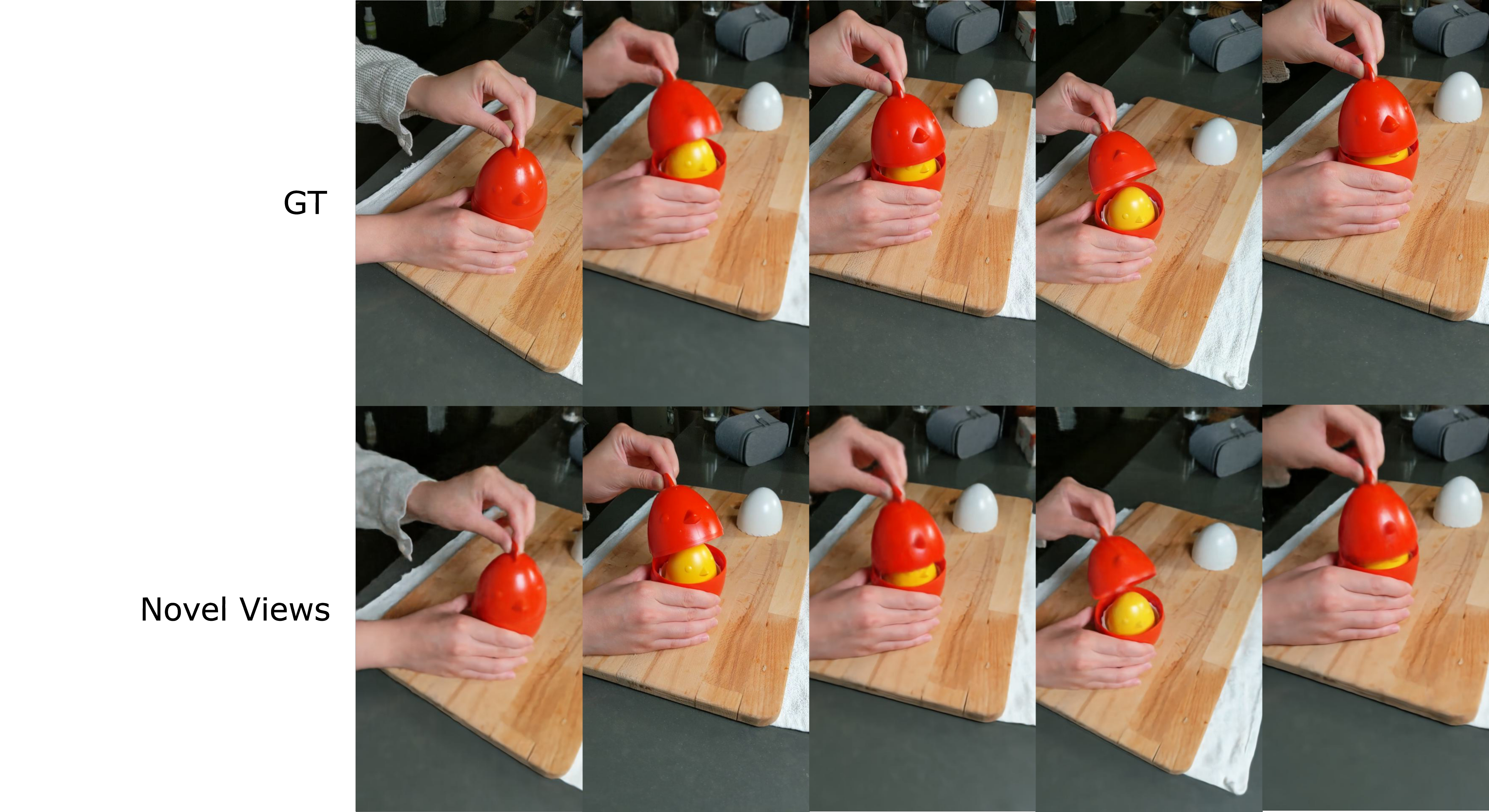}
\caption{Interpolated novel views on the HyperNeRF Chicken sequence.}
\label{fig:chicken}
\end{figure}

\end{document}